\documentclass[letterpaper, 10 pt, conference]{ieeeconf}  

\IEEEoverridecommandlockouts                              

\usepackage{cite}
\usepackage{amsmath} 
\usepackage{amssymb}  
\usepackage[dvipsnames]{xcolor}
\usepackage{tikz}
\usetikzlibrary{bayesnet}
\usepackage{graphicx}
\usepackage{caption}
\usepackage{subcaption}
\usepackage{booktabs}   
\usepackage{multirow}

\let \labelindent \relax
\usepackage{enumitem}

\usepackage{algorithm}
\usepackage{algpseudocode}
\usepackage[linesnumbered,vlined,noend,algo2e]{algorithm2e}

\newcommand{\graspHeuristic}{
\begin{algorithm2e}[H]
\SetAlgoLined
\SetKwFunction{FGrasp}{pickHeuristic}
  \SetKwProg{Fn}{}{:}{}
  \Fn{\FGrasp{robotBase $b$, object $o$}}{
        \lIf{reachable($b$, $o$)}{
            \Return 0
        }
        \lElse{
            \Return $\infty$
        }
  }
\end{algorithm2e}}

\newcommand{\placeHeuristic}{
\begin{algorithm2e}[H]
\SetKwFunction{FPlace}{placeHeuristic}
  \SetKwProg{Fn}{}{:}{}
  \Fn{\FPlace{robotBase $b$, object $o$, target $t$}}{
    \eIf{reachable($b$, $t$)}{
        \lIf{$t$ == goalTray}{
            \Return -1
        }
        \lElse{
            \Return 0
        }
    }{
        \Return $\infty$
    }
  }
\end{algorithm2e}}

\newcommand{\stepHeuristic}{
\begin{algorithm2e}[H]
\SetKwFunction{FStep}{stepHeuristic}
  \SetKwProg{Fn}{}{:}{}
  \Fn{\FStep{robotBase $b$, stepTarget $t$, navigationTarget $g$, chainLen $n$}}{
    \eIf{reachable($b$, $t$)}{
        stepSize $\gets$ dist($t$,$g$) - dist($b$,$g$)\;
        \Return stepSize\;
    }{
        \Return $\infty$\;
    }
  }
\end{algorithm2e}}

\newcommand{\connectHeuristic}{
\begin{algorithm2e}[H]
\SetKwFunction{FHeuristic}{connectHeuristic}
  \SetKwProg{Fn}{}{:}{}
  \Fn{\FHeuristic{priorDecision $d$, effectorA $a$, effectorB $b$}}{
    \lIf{reachable($a$,$b$) $\land$ $d~\neq$ disconnect}{
        \Return 0
    }
    \lElse{
        \Return $\infty$
    }
  }
\end{algorithm2e}}

\newcommand{\disconnectHeuristic}{
\begin{algorithm2e}[H]
\SetKwFunction{FHeuristic}{disconnectHeuristic}
  \SetKwProg{Fn}{}{:}{}
  \Fn{\FHeuristic{priorDecision $d$}}{
    \lIf{$d~\neq$ connect}{
        \Return 0
    }
    \lElse{
        \Return $\infty$
    }
  }
\end{algorithm2e}}
\usepackage{float}
\newfloat{algorithm}{t}{lop}

\makeatletter
\newcommand{\removelatexerror}{\let\@latex@error\@gobble}
\makeatother

\usepackage{hyperref}
\usepackage[capitalise]{cleveref}

\usepackage{wrapfig}
\usepackage{booktabs}

\usepackage{tabularx,booktabs} 
\usepackage{eqparbox}

\newcommand{\PP}{\mathcal{P}}

\newcommand{\XX}{\mathcal{X}}
\newcommand{\R}{\mathbb{R}}
\newcommand{\N}{\mathbb{N}}
\renewcommand{\AA}{\mathcal{A}}
\renewcommand{\SS}{\mathcal{S}}

\newcommand{\st}{\quad\text{s.t.}\quad}

\title{\LARGE \bf 
 RHH-LGP: Receding Horizon And Heuristics-Based Logic-Geometric Programming For Task And Motion Planning \\
 }

\author{Cornelius V. Braun$^{1, 4}$, Joaquim Ortiz-Haro$^{1,3}$, Marc Toussaint$^{1}$, Ozgur S. Oguz$^{1,2}$
\thanks{$^{1}$Learning and Intelligent Systems Group, TU Berlin, Germany}%
\thanks{$^{2}$Bilkent University, Turkey
        }%
\thanks{$^{3}$International Max Planck Research School for Intelligent Systems (IMPRS-IS)}%
\thanks{$^{4}$Imperial College, London, United Kingdom}%
\thanks{\textbf{Note:} This work has been submitted to the IEEE for possible publication. Copyright may be transferred without notice, after which this version may no longer be accessible.}%
}

\begin{document}

\maketitle
\thispagestyle{empty}
\pagestyle{empty}

\begin{abstract}
Sequential decision-making and motion planning for robotic manipulation induce combinatorial complexity. 
For long-horizon tasks, especially when the environment comprises many objects that can be interacted with, planning efficiency becomes even more important. 
To plan such long-horizon tasks, we present the RHH-LGP algorithm for combined task and motion planning (TAMP).
First, we propose a TAMP approach (based on Logic-Geometric Programming) that effectively uses geometry-based heuristics for solving long-horizon manipulation tasks.
The efficiency of this planner is then further improved by a receding horizon formulation, resulting in RHH-LGP. 
We demonstrate the robustness and effectiveness of our approach on a diverse range of long-horizon tasks that require reasoning about interactions with a large number of objects.
Using our framework, we can solve tasks that require multiple robots, including a mobile robot and snake-like walking robots, to form novel heterogeneous kinematic structures autonomously.
By combining geometry-based heuristics with iterative planning, our approach brings an order-of-magnitude reduction of planning time in all investigated problems.







\end{abstract}

\section{Introduction}\label{secIntroduction}
Combined task and motion planning (TAMP) breaks down complex sequential manipulation problems into two separate processes: high-level decision-making and motion planning~\cite{dornhege2009integrating,dantam16incremental,lagriffoul16combining,hadfield-menell16sequential,ferrer-mestres17combined,gaschler18kaboum,shoukry18smc,colledanchise19towards,garrett2021integrated}.
A high-level decision (or action) is represented by symbolic abstractions that describe its pre- and post-conditions.
Given a goal (usually symbolic), a decision sequence is first found, and then a feasible motion trajectory is computed for the robot(s) to realize the given task.
However, when the task requires executing a long sequence of high-level actions, and there is a large set of objects to consider in the environment, the combinatorial complexity of those problems makes them computationally intractable.
This is exacerbated when a team of robots has to carry out such tasks together.

In essence, developing an efficient TAMP solver for long-horizon tasks remains a challenging problem~\cite{nair2019hierarchical,driess2021learning}.
There are two main approaches: (\textit{i}) integrating heuristics to reduce the symbolic search space, and (\textit{ii}) decomposing the problem into solvable subtasks.
Prior studies on TAMP use heuristics~\cite{erdem11combining,srivastava2013using,srivastava2014combined,erdem15ingtegrating}, some of them leveraging geometric information acquired from the environment to improve planning performance~\cite{cambon09hybrid,garrett2015ffrob}.
However, these methods rely on calling the motion planner to compute heuristic cost values, which deteriorates their performance as the required action length increases.
In~\cite{vega2018admissible}, geometric abstractions, which do not require using the actual motion planner, are utilized for effective task planning. 
However, the motion planning problem is simplified since the tasks require 2D navigation in maze-like environments.
In~\cite{akbari2019combined}, a relaxed geometric reasoning approach enables pruning geometrically infeasible actions, and in~\cite{wells2019learning} a Support Vector Machine is used to estimate the feasibility of actions.
However, most of the previous works that used heuristics for TAMP have focused on smaller problem instances, which either required shorter action sequences to be solved~\cite{akbari2019combined, driess2020deep,wells2019learning} or contained few objects in the environment~\cite{akbari2016task}.
\begin{figure}[t]
    \centering
    \begin{subfigure}[b]{.2\textwidth}
        \centering
        \includegraphics[width=\linewidth, trim=25cm 1cm 25cm 5cm,clip]{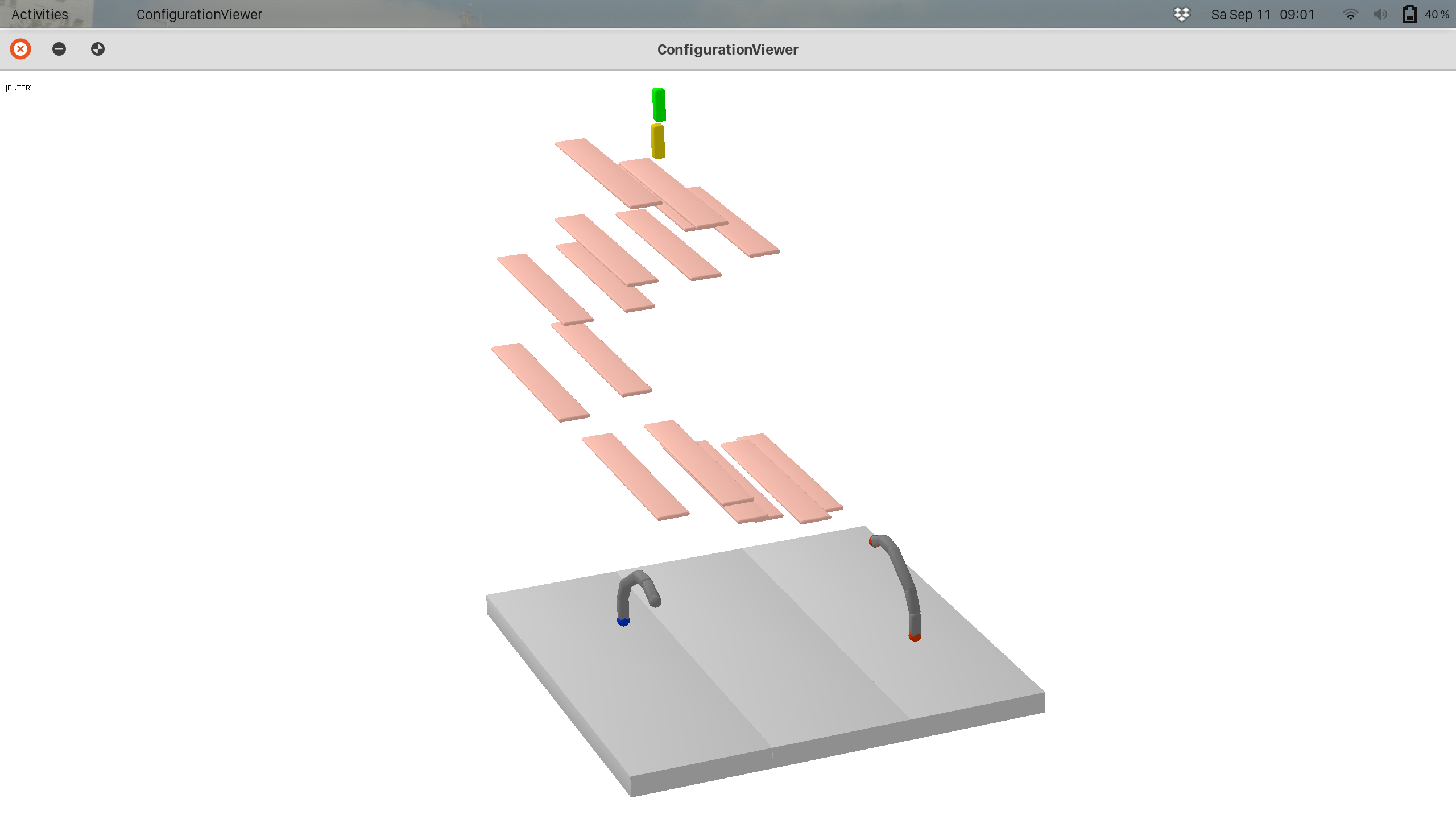}
        \caption{}\label{fig:complex}
    \end{subfigure}
    \begin{subfigure}[b]{.2\textwidth}
        \includegraphics[width=\linewidth,trim=19cm 1cm 19cm 5cm,clip]{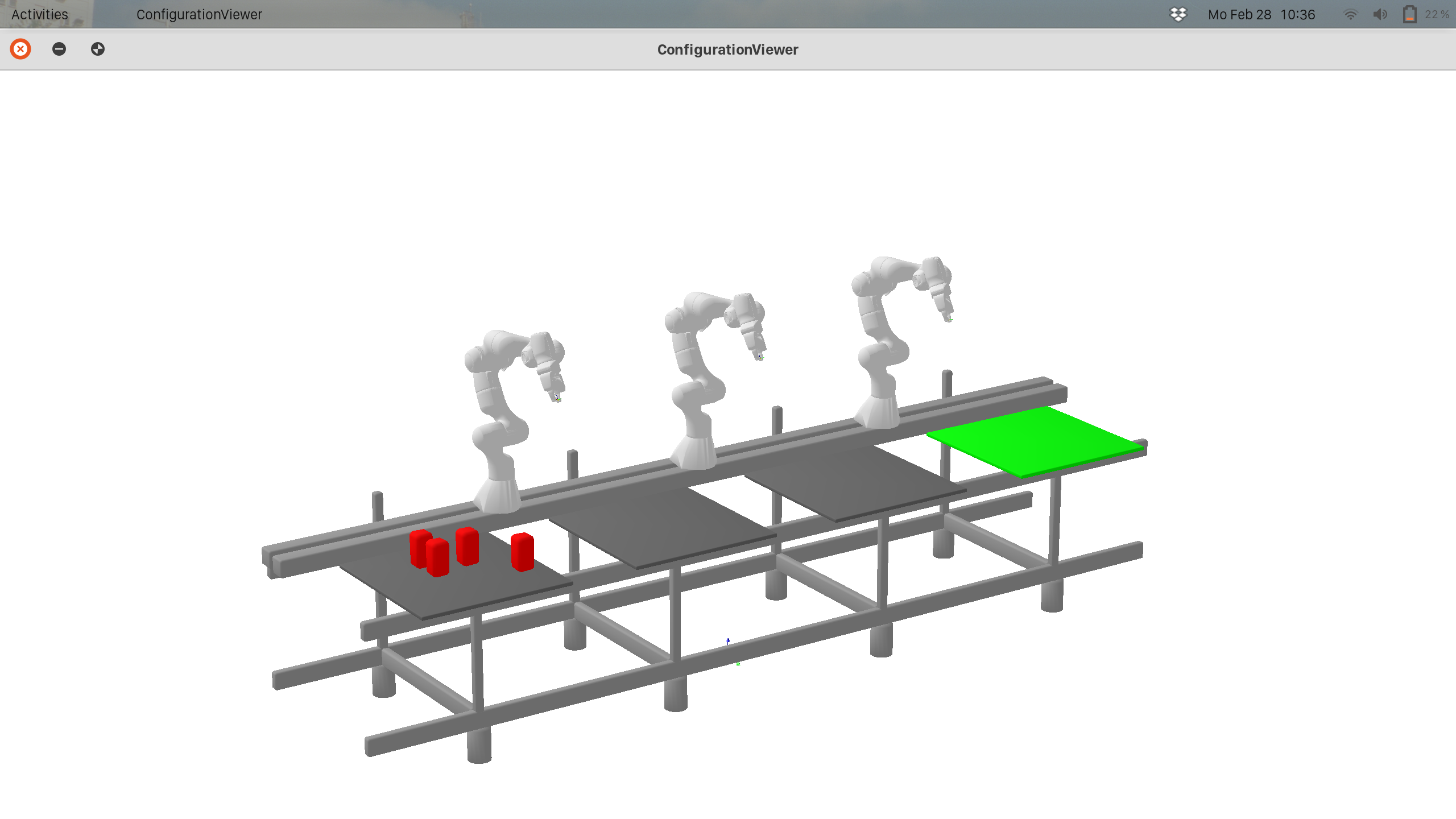}
        \caption{}\label{fig:else}
    \end{subfigure}
    \begin{subfigure}[b]{.2\textwidth}
        \includegraphics[width=\linewidth,trim=25cm .75cm 25cm 25cm,clip]{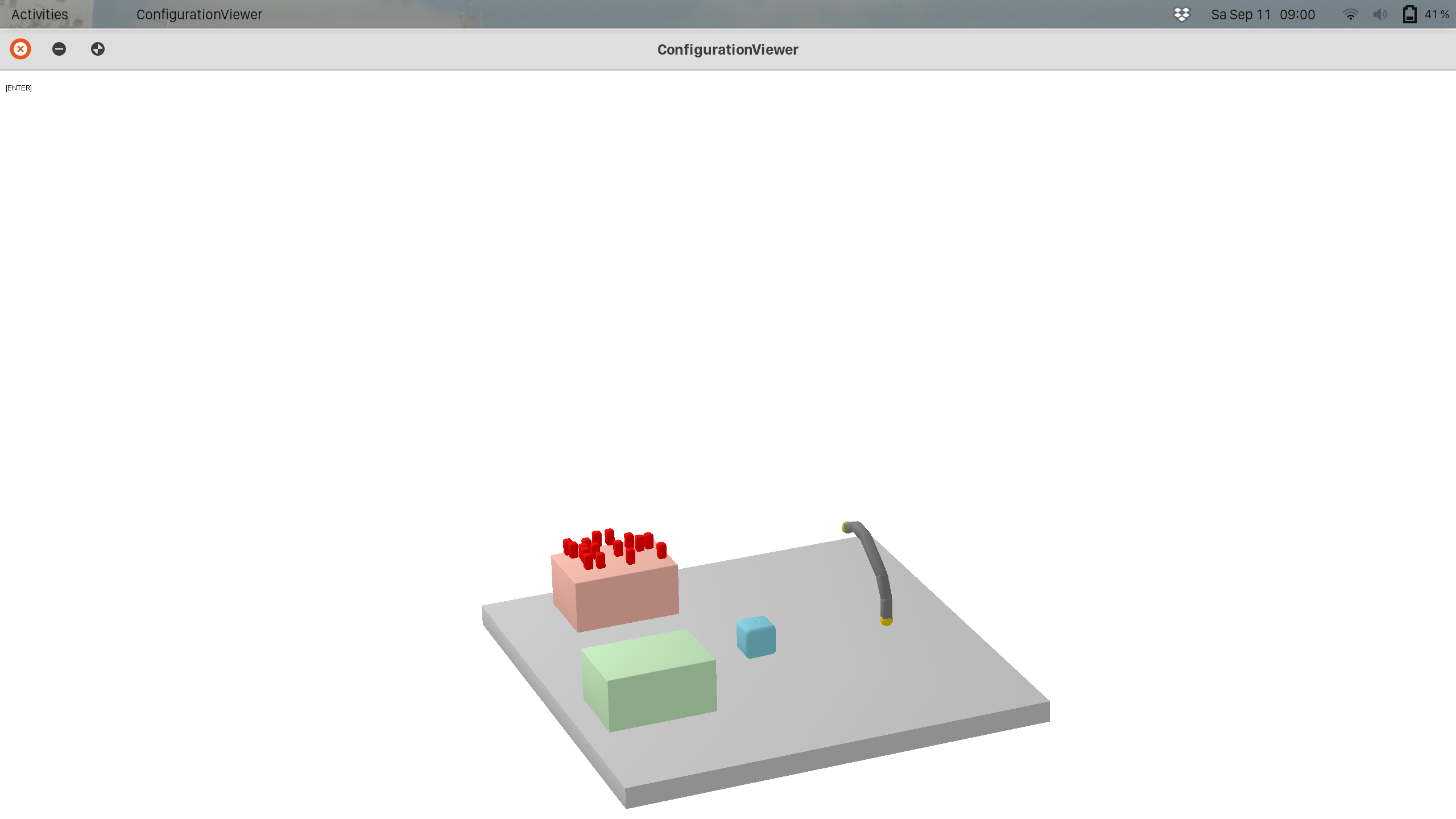}
        \caption{}\label{fig:mobileManip}
    \end{subfigure}
    \begin{subfigure}[b]{.2\textwidth}
        \includegraphics[width=\linewidth,trim=20cm 1cm 20cm 25cm,clip]{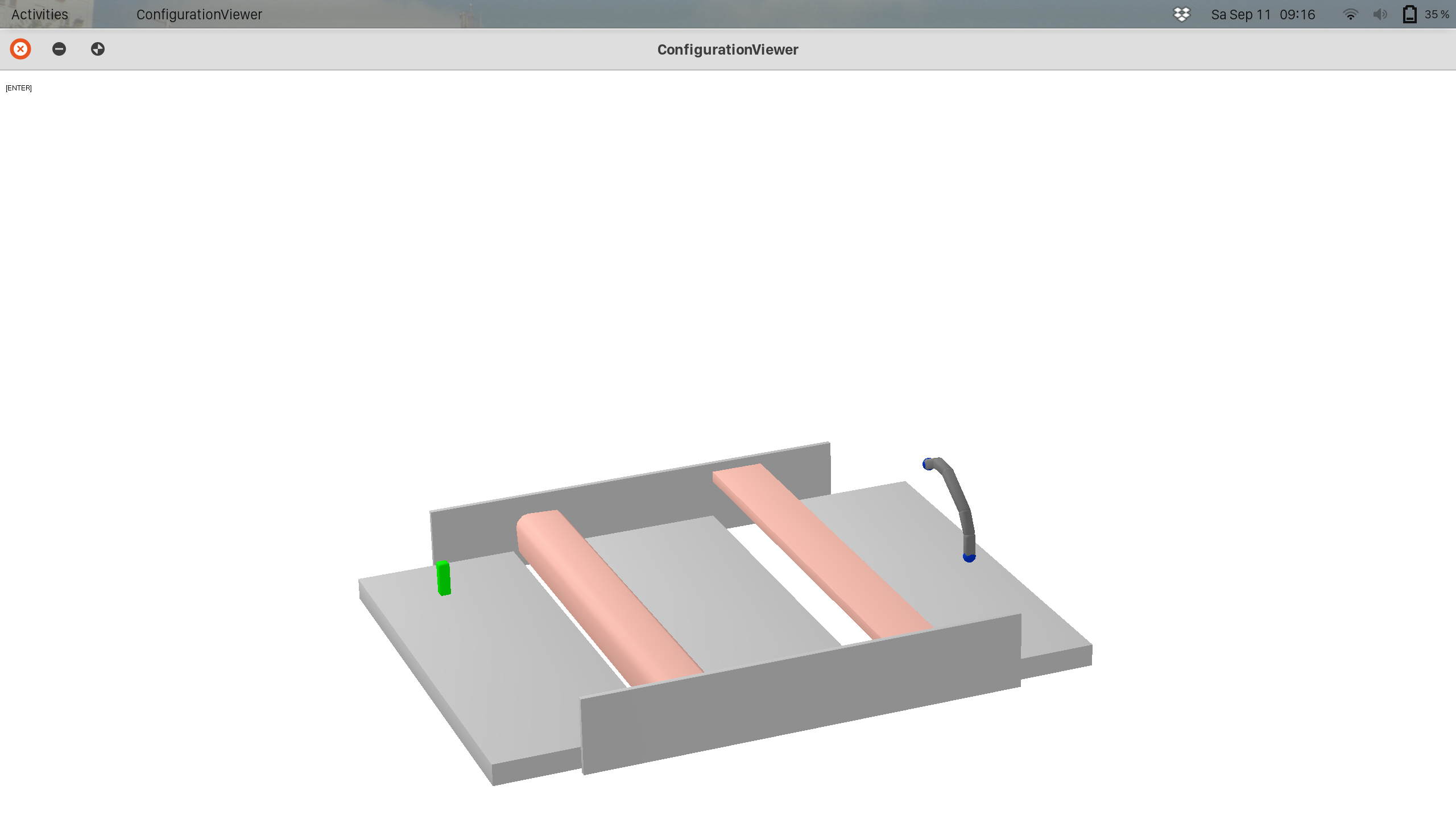}
        \caption{}\label{fig:obs}
    \end{subfigure} 
    \caption{
    Benchmark problems.
    (\textit{a}) Climbing task with multiple crawler robots and multiple goals, i.e. picking both boxes (green and yellow). 
    (\textit{b}) Pick-and-place task with multiple robotic arms and the goal to transfer several objects to the green target.
    (\textit{c}) A long-horizon pick-and-place task, where a mobile (blue) and a crawler (gray) robot have to dynamically 'connect' to accomplish the task, i.e. placing each cup on the green table.
    (\textit{d}) Obstacle avoidance task with the goal for one crawler robot to reach the green box.
    }
    \label{fig:problems}
\end{figure}
Another technique to handle the challenges of long-horizon TAMP is iterative planning or problem decomposition.
Such planners plan and/or execute only a portion of the task at each step.
The benefit of this approach is that large-scale problems are partitioned into computationally tractable subproblems.
Previous works have tackled these tasks by defining several subgoals that have to be fulfilled to complete an entire long-horizon task~\cite{driess2019hierarchical,nair2019hierarchical, gupta2019relay}, or receding horizon approaches are adopted, where in each planning iteration, an action sequence of fixed length is planned~\cite{hartmann2020robust}, and only a subset of those actions of each plan is executed~\cite{castaman2020receding, garrett2020online}.
The planner in~\cite{hartmann2020robust} relies on task-specific subgoal definitions~\cite{hartmann2020robust}, while the approach in~\cite{castaman2020receding} focuses on single task domains like pick-and-place tasks, and in~\cite{garrett2020online} the planner addresses partially observable environments and thus puts focus on information gain during planning.
While these methods improve the efficiency of solving sequential manipulation problems either on singular problems or for a single robot, an integrated TAMP approach that can be effectively applied on long-horizon settings from different domains, including navigation tasks, pick-and-place scenarios, and the dynamic formation of modular robots, is still missing.


We tackle the problem of solving long-horizon robotic manipulation tasks by proposing a heuristics augmented TAMP approach.
Our method builds on logic-geometric programming (LGP), which formalizes the relation between discrete decision variables and a continuous nonlinear mathematical program (NLP) that represents the underlying continuous world problem~\cite{toussaint2015logic,toussaint2017multi,de2020learning}.
However, the prior LGP approaches have the drawback of exploring geometrically infeasible action sequences, since the symbolic search over discrete actions ignores the geometrical constraints of the scene. 
In addition, solving the entire problem at once is computationally difficult not only due to the combinatorial complexity of tasks involving many objects but also because motion planning of a long sequence of actions requires solving an NLP with a large set of kinematic constraints.

To address these challenges, this work introduces the following contributions
\begin{itemize}
    \item a novel geometry-based heuristics integrated LGP approach, and
    \item a receding horizon formulation of LGP that allows solving long-horizon TAMP problems effectively, while
    \item enabling the generation of cooperative and modular multi-agent behavior.
\end{itemize}
We demonstrate the flexibility and effectiveness of the proposed approach in problem settings from different domains that require solving complex long-horizon tasks (Fig.~\ref{fig:problems}).
These tasks include a multi-robot pick-and-place task (Fig.\ \ref{fig:else}), a scenario that requires climbing a large set of stairs (up to 64) by snake-like locomoting robots (\textit{crawlers}, Fig.~\ref{fig:complex}), 
rearrangement tasks in which heterogeneous robots have to autonomously form novel kinematic structures to carry out manipulation actions (Fig.~\ref{fig:mobileManip}),
and a complex \textit{parkour} setting where the crawler has to find ingenious movement strategies to reach the goal while avoiding obstacles (Fig.~\ref{fig:obs}).

\section{Background}\label{secBackground}

\subsection{Logic Geometric Programming}
This work is based on Logic Geometric Programming (LGP),\cite{ Toussaint-RSS-18, toussaint2017multi} as the underlying framework for solving TAMP problems.  We will now briefly outline the optimization formulation of LGP that is given by eq. \eqref{eq:LGP}.

Let $\XX \subset \R^n \times SE(3)^m$ be the configuration space of $m$ rigid objects and $n$ articulated joints with initial condition $x_0$. The goal of LGP is to optimize a sequence of symbolic actions $a_{1:K}$ and states $s_{1:K}$ and the corresponding continuous path in configuration space $x(t): t \in \mathbb{R} \to \mathcal{X}$ to achieve a symbolic goal $g$.
Positions, velocities and accelerations are denoted with $\bar{x}=(x, \dot{x}, \ddot{x})$. The space of symbolic states $s \in \mathcal{S}$ and actions $ a \in \mathcal{A}(s)$ is discrete and finite, and the transitions $s_{k-1} \to s_k$ are determined by a first-order logic language (similar to PDDL).%

\begin{figure}[!htbp]
\begin{subequations}\label{eq:LGP}
\begin{align}
  \min_{x,s_{1:K},a_{1:K},K} & \int_0^{KT} c( \bar{x}(t) , s_{k(t)}) ~ dt \\
  \st \nonumber 
& \bar{x}(0) = x_0, \\
 \forall{t \in [0,KT]}:~ 
&  h_{path} ( \bar{x}(t), s_{k(t)}) = 0 \label{subeq:1c}\\ 
& g_{path} ( \bar{x}(t), s_{k(t)}) \leq 0 \label{subeq:1d}\\ 
  \forall{k \in 1,...,K}:~
& h_{switch} ( \bar{x}(t_k),a_k) = 0 \label{subeq:1e}\\ 
& g_{switch} ( \bar{x}(t_k), a_k) \leq 0 \label{subeq:1f}\\ 
& a_k \in \AA(s_{k-1}) \\
& s_k \in succ(s_{k-1}, a_k) \\
& s_K \in \SS_{goal}(g)
\end{align}
\end{subequations}
\end{figure}
For a fixed action and state sequence $a_{1:K}$ and $s_{1:K}$, the functions $h(\cdot),g(\cdot),c(\cdot)$ are continuous and piecewise differentiable, and the optimization of the path is a nonlinear program (NLP). Therefore, the main idea behind LGP is to combine a discrete search on the symbolic level (i.e., find a sequence of symbolic states that leads to the goal) with joint nonlinear optimization to compute paths that fulfill the constraints and to assert whether a symbolic action sequence admits a continuous realization.

\subsection{Multi-Bound Tree Search}
The discrete variables of the LGP formulation in eq.\ \eqref{eq:LGP} induce a decision tree that contains sequences of symbolic states, starting from $s_0$. The leaf nodes where $s \in \mathcal{S}_{\text{goal}}(g)$  are potential candidates for a solution. Each node can be tested for feasibility by solving the NLP induced by the sequence of states from the root to the current node.
%
However, solving the continuous path problem is expensive and the number of candidate NLPs is generally too high. To alleviate this issue, Multi-Bound Tree Search (MBTS) \cite{toussaint2017multi} first solves relaxed versions of eq.~\eqref{eq:LGP}. The feasibility of each relaxed problem is a necessary condition for the feasibility of the original NLP, i.e., acting as lower bounds, whilst being computationally simpler.
In this work, we use the two bounds: $\PP_{seq}$ (joint optimization of the mode-switches of the sequence),  and $\PP_{path}$ (full problem).

\section{Informed Tree Search with Novel Geometric Heuristics}\label{secTamp}
\subsection{Limitations of standard MBTS for solving LGP}
The running time of optimization-based TAMP planners is dominated by the number of calls to the nonlinear optimizer that evaluates the geometric and kinematic feasibility of a symbolic plan \cite{toussaint2015logic,zhao2021sydebo,wells2019learning}.
In MBTS, the interleaving between symbolic search and continuous optimization of paths can be achieved in two different ways: either by (\textit{i}) solving NLPs only for the sequences of symbolic actions that lead to the goal or (\textit{ii}) solving NLPs of intermediate subsequences to guide the symbolic search. Both approaches are inefficient in large symbolic spaces. 
With approach (\textit{i}), the goal-reaching sequences have never been informed about the geometry and most of the candidate action sequences are infeasible, while in approach (\textit{ii}) most of the computed NLPs correspond to subsequences that do not lead to the goal.

\subsection{Heuristics-enriched LGP}
In this work, we propose a heuristic-based formulation that guides the symbolic search and prunes geometrically infeasible solutions. Our heuristics are based on fast geometry checks and are an order-of-magnitude faster than solving intermediate NLPs.
The heuristic-based LGP involves two components: \textit{action-specific heuristics}, which incorporate geometric information to estimate the costs of performing a given action in a certain state, and the \textit{interface for heuristics} which integrates the action-specific heuristics into the LGP planning algorithm.

\paragraph{An interface for heuristics} 
The optimal (but intractable) cost-to-go of a node $s_i$ in the LGP-tree is defined as the cost of the LGP problem conditioned on the current sequence of symbolic states (the path from the root $s_0$ to the node), i.e. solving eq.\ \eqref{eq:LGP} with fixed $s_0, s_1 , ... , s_i$, taking infinite value if the conditioned LGP problem is infeasible. 

To inform symbolic search, we propose an LGP-interface for heuristics that uses a cost-to-go function, which assigns a \textit{heuristic} cost-to-go value to each node expanded in the search tree.
This heuristic value can be computed from the current symbolic state $s_i$, from the previous symbolic action (i.e., transition $s_{i-1} \to s_{i}$) or from the whole subsequence starting at the root $s_0,\ldots, s_i$. 
To label a node of the tree as infeasible, the cost-to-go function assigns an infinite cost that prevents future expansion.
We define a problem-specific cost-to-go function upon initialization of the LGP problem.
For example, we define the cost-to-go function that assigns an infinite cost-to-go value to any action sequence that includes a robot picking up an object which is out of reach (see details below).
The new cost-to-go is used to inform the multi-bound tree search in two steps: choosing which node to expand next and choosing which NLP to optimize next. 

The complete RHH-LGP algorithm is shown in Alg.~\ref{alg:lgp}.
Multi-bound tree search with geometric heuristics corresponds to the inner loop (with our contribution highlighted). 
\RestyleAlgo{ruled}

\begin{algorithm2e}[!htb]
    \footnotesize
    \DontPrintSemicolon
    \SetAlgoLined
    \SetInd{0.5em}{0.5em}
    
    \SetKwProg{Fn}{Function}{:}{}
    \SetKwProg{Alg}{Alg}{:}{}
    \SetKwFunction{FOpt}{evaluateBound}
    \SetKwFunction{FAlg}{RHLGP}
    \SetKwFor{Until}{until}{do}{end}
    
    \SetKwInOut{Input}{input}
    \SetKwInOut{Output}{output}
    
    \Input{symbolic goal $g$, initial symbolic state $s_0$, \\
    initial kin. state $x_0$, horizon $h$}
    
    \vspace{2pt}

    $s \gets s_0$ \;
    \Until{$s \in \mathcal{S}_{goal}(g)$}{
       PathFound $ \gets $ false \;
          SymbolicSearchQueue $ \gets \{ s \} $ \;
        SequenceNLPQueue $\gets ~ \emptyset$ \;
        PathNLPQueue $\gets ~ \emptyset$ \;
        \Until{  PathFound }{
                \textit{// Tree Search Symbolic Level}\;
                {\color{RubineRed} node $\gets$ SymbolicSearchQueue.argmin(heuristicCost)\; }
                \eIf{node $\in \mathcal{S}_{goal}(g)$ }
                {
                                SequenceNLPQueue.append(node)
                }
                {
                  SymbolicStateQueue.append(node.expand())

                }
                \textit{// evaluate bounds of best candidates}\;
                {\color{RubineRed} 
                candidateNode $ \gets $ SequenceNLPQueue.argmin(heuristicCost) }\\
                {\color{RoyalBlue} skeleton $\gets$ candidateNode.actions[$0:h$] }\;
                 \If {
             solveSequenceNLP(skeleton)
                 }   
                 {
                 PathNLPQueue.append(candidateNode) 
                 }

                    candidateNode $ \gets $ PathNLPQueue.argmin(sequenceCost) \\
                {\color{RoyalBlue} skeleton $\gets$ candidateNode.actions[$0:h$] }\;
                          \If {
             solvePathNLP(skeleton)
                 } 
                 {
                 PathFound $ \gets $ true\;
                 s $ \gets $ candidateNode\;
                 }
                 
         }
     }

\caption{RHH-LGP algorithm, with heuristics-related contributions in pink and receding horizon-related contributions in blue.}\label{alg:lgp}
\end{algorithm2e}
In short, the multi-bound tree search uses three queues: \texttt{SymbolicSearchQueue} (nodes discovered during tree search); \texttt{SequenceNLPQueue} (candidate nodes for solving the bound sequence NLP); \texttt{PathNLPQueue} (candidate nodes for solving the full path NLP).
The tree search performs the following steps iteratively until a path is found: new nodes are discovered, nodes that reach the symbolic goal are moved to the queue of candidates for sequence NLP ($\PP_{seq}$) optimization, and nodes with a feasible sequence NLP are moved to the queue of candidates to solve the path NLP ($\PP_{path}$). 

\paragraph{Geometry-based reachability check}
\begin{figure}
    \centering
    \begin{subfigure}[b]{.13\textwidth}
        \centering
        \includegraphics[scale=.35,trim=8cm 1cm 11cm 2cm,clip]{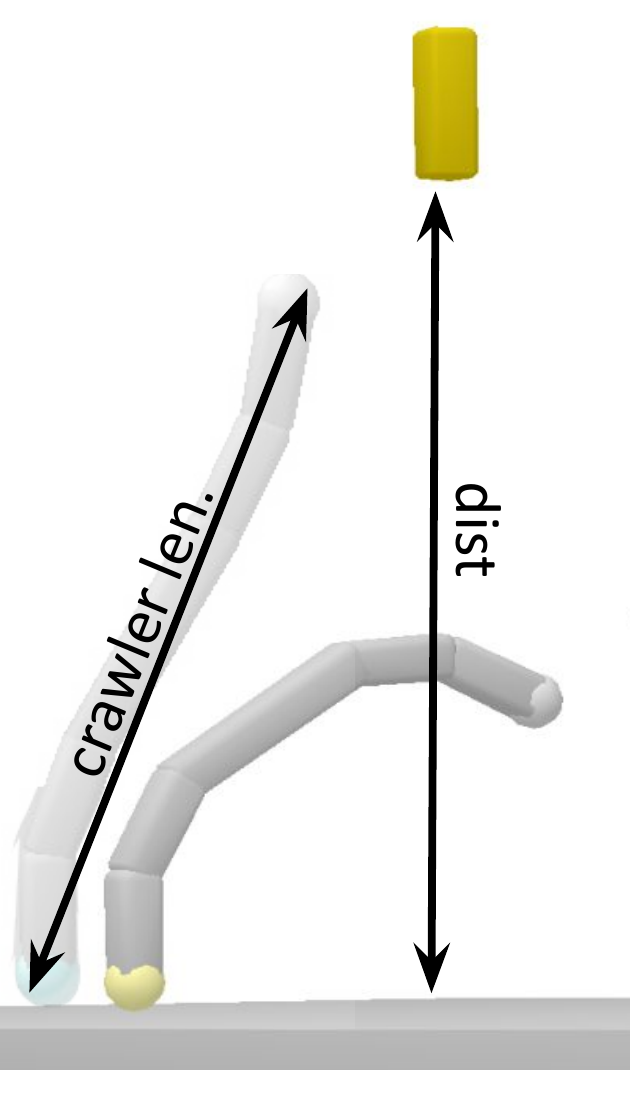}
        \caption{pick heuristic} 
        \label{fig:hGrasp}
    \end{subfigure}
    \begin{subfigure}[b]{.15\textwidth}
        \includegraphics[scale=.35,trim=9cm 3cm 9cm 1cm,clip]{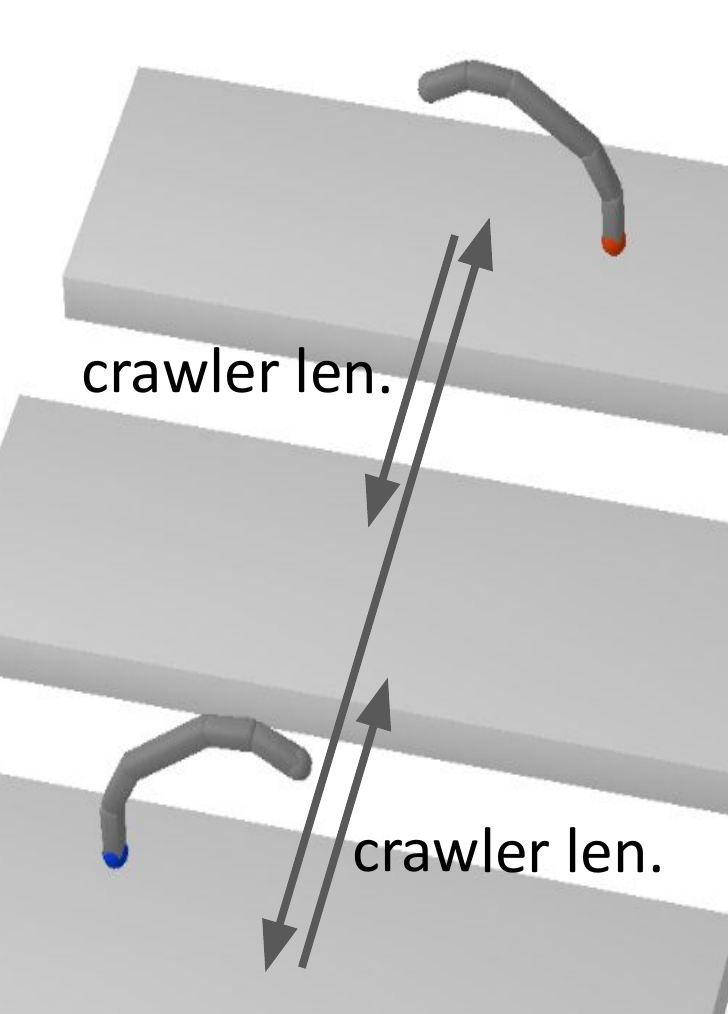}
        \caption{connect heuristic}
        \label{fig:hConnect}
    \end{subfigure}
    \begin{subfigure}[b]{.15\textwidth}
        \includegraphics[scale=.35,trim=8.5cm 3cm 0 2cm,clip]{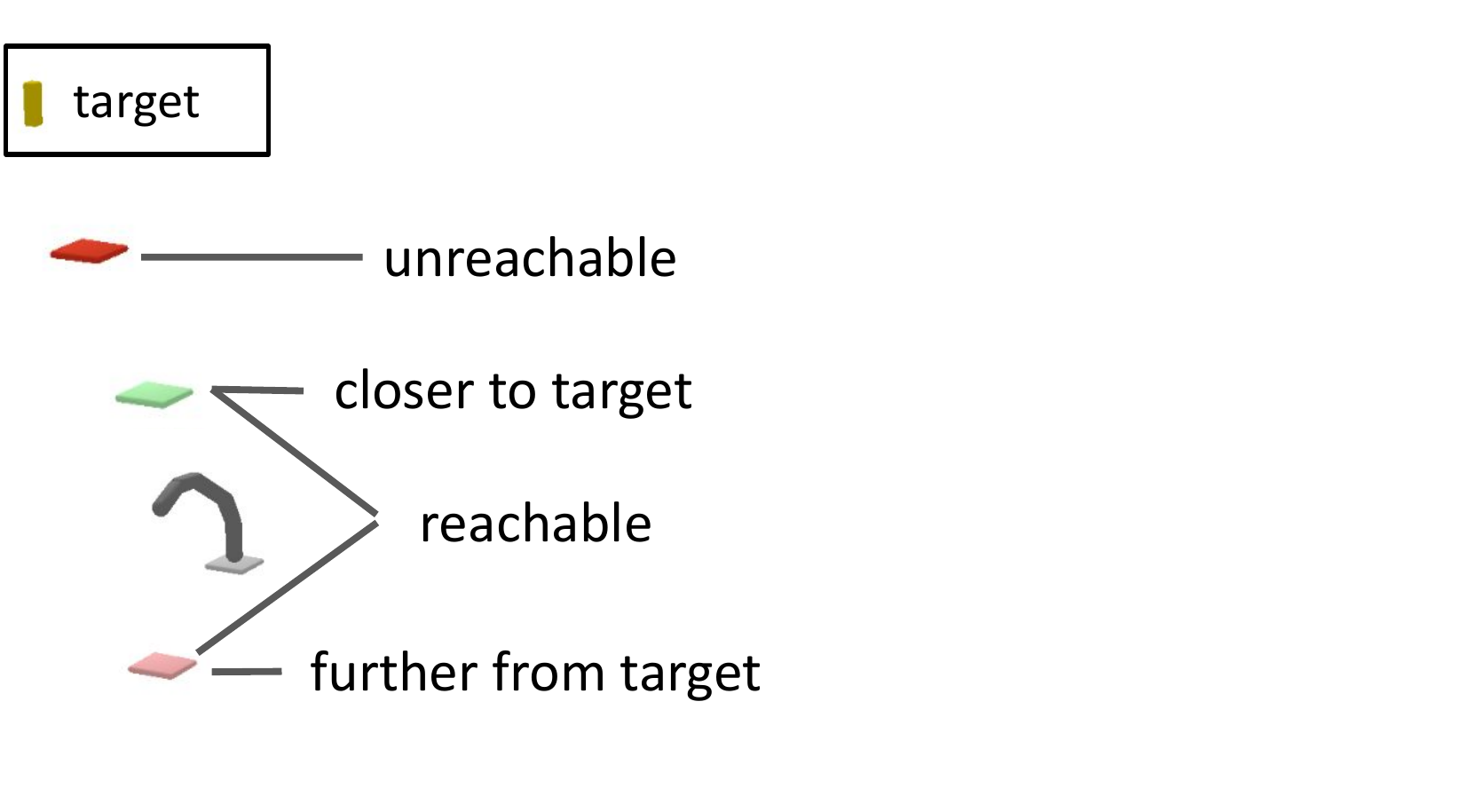}
        \caption{step heuristic}
        \label{fig:hStep}
    \end{subfigure}
    \caption{A subset of the action-specific heuristics we present.}
    \label{fig:reachability}
\end{figure}

The heuristic cost-to-go is computed based on the symbolic actions and states, but the underlying geometric functions require a kinematic configuration.
Since calling an NLP solver to compute such configuration would be too expensive, the geometry-based reachability checks are performed using the initial kinematic configuration of the continuous state space.
The reachability checks that we use compare the distances between objects with the reach of the robots, which in our case is defined by the length of the crawler or robotic arm.
This strategy can be applied when the objects or target surfaces in the environment are immobile, e.g., stairs, floor, and tables.
The idea is illustrated in Fig. \ref{fig:hGrasp}, where the distance of the target to the base is longer than the length of the extended crawler, from which it follows that picking the object immediately will be infeasible.

\subsection{Heuristics for Solving Long Horizon Tasks}
In this section, we present \textit{action-specific heuristics} that help to formulate cost-to-go functions for solving long locomotion and manipulation tasks with three types of (modular) robots: the \textit{crawler}, the \textit{mobile-base} and a \textit{robotic arm}:

\begin{enumerate}[label=\textit{(\roman*)}]
  \item \textit{crawler}: snake-like robot, composed
    of 7 joints and two end-effectors. The end-effectors are spherical and can be used to provide stable support for walking and for touch-based picking and placing.
    In addition, two crawler modules can connect at their end-effectors to form one large crawler.
    
   \item \textit{mobile-base}: cube-shaped robot, able to move freely in the xy-plane.
   The crawler robots can connect to the mobile base in order to form a mobile manipulator robot that can in turn pick and place objects.
   
   \item \textit{Robotic arm}: the 7 DoF (Franka Panda) robot attached to a surface. It can pick and place objects.
\end{enumerate}

Given a symbolic action, an action-specific heuristic function returns a cost-to-go value that approximates the cost to reach the goal after executing that action (with an infinite value if the action is geometrically infeasible). We present the following heuristics (Table~\ref{tbl:heuristics}) that can be combined as building blocks to create novel cost-to-go functions.

The \texttt{pickHeuristic} checks if the object to be picked is within reach (Fig.~\ref{fig:hGrasp}).
To this end, the heuristic checks if the target object is within reach of the robot that stands on the given base object $b$.

Similarly, the \texttt{placeHeuristic} checks if the target surface for placing is within reach. Additionally, it incentives placing objects in the goal region.

The \texttt{stepHeuristic} checks if the  target object $t$ can be reached from the object $b$, which the robot is standing on before the step (Fig.~\ref{fig:hStep}).
If the step is feasible, the heuristic returns the amount by which the distance of the robot to its goal of navigation will be reduced by the step's size, effectively providing guidance for navigation planning.

The \texttt{connectHeuristic} checks if the two robot modules that are to connect are within reach of each other (Fig.~\ref{fig:hConnect}).
Furthermore, the heuristic prunes decision sequences that contain contiguous disconnect and disconnect actions.
Similarly, the \texttt{disconnectHeuristic} prevents the disconnection of modular robots that have been connected in the prior action, as such a decision sequence will never be optimal. 
\RestyleAlgo{plain}
\begin{table}[tbp]
\removelatexerror
\resizebox{\linewidth}{!}{
\begin{tabular}{l|l}
    \multicolumn{1}{c|}{\textbf{Action}}      & \multicolumn{1}{c}{\textbf{Heuristic}} \\[0.5em] \hline 
    pick                   & \graspHeuristic       \\ \hline 
    place                   & \placeHeuristic       \\ \hline 
    \shortstack{step \& \\ stepTogether}   & \stepHeuristic        \\ \hline 
    connect                 & \connectHeuristic     \\ \hline 
    disconnect              & \disconnectHeuristic  \\
\end{tabular}}
\caption{Action specific heuristics used in the present work.}\label{tbl:heuristics}
\addcontentsline{toc}{subsection}{\nameref{tbl:heuristics}}
\end{table}

\subsection{Completeness of Heuristics-enriched MBTS}
An algorithm is complete if it finds a solution for a given feasible problem.
MBTS performs search over the set of NLPs that correspond to all possible action sequences that lead to the symbolic goal. 
Under the assumption that a feasible action sequence corresponds to at least one feasible NLP (i.e., the motion optimization does find an optimum if such a solution exists), the MBTS algorithm is complete~\cite{driess2021learning}.
In the case of \texttt{pick}, \texttt{place} and \texttt{step} actions, the geometry-based heuristics only prune action sequences from the search tree that are infeasible, e.g.\ they contain actions such as grasping an object that is out of reach.
In this case, the completeness of the MBTS algorithm still holds.
In addition, the heuristics prune action sequences that contain subsequent \texttt{connect} and \texttt{disconnect} actions.
In this case, there always exists another feasible action sequence without these two consecutive actions, that leads to the same symbolic goal state.
Therefore, the completeness of the MBTS algorithm is still preserved in this case.
Thus it follows that the heuristic-enriched MBTS algorithm is complete.

\section{Receding horizon-LGP}
\begin{figure*}[t!]
\centering
\begin{subfigure}[b]{.17\textwidth}
  \centering
  \includegraphics[width=\textwidth,height=\textwidth,trim=10cm 0cm 10cm 2.75cm,clip]{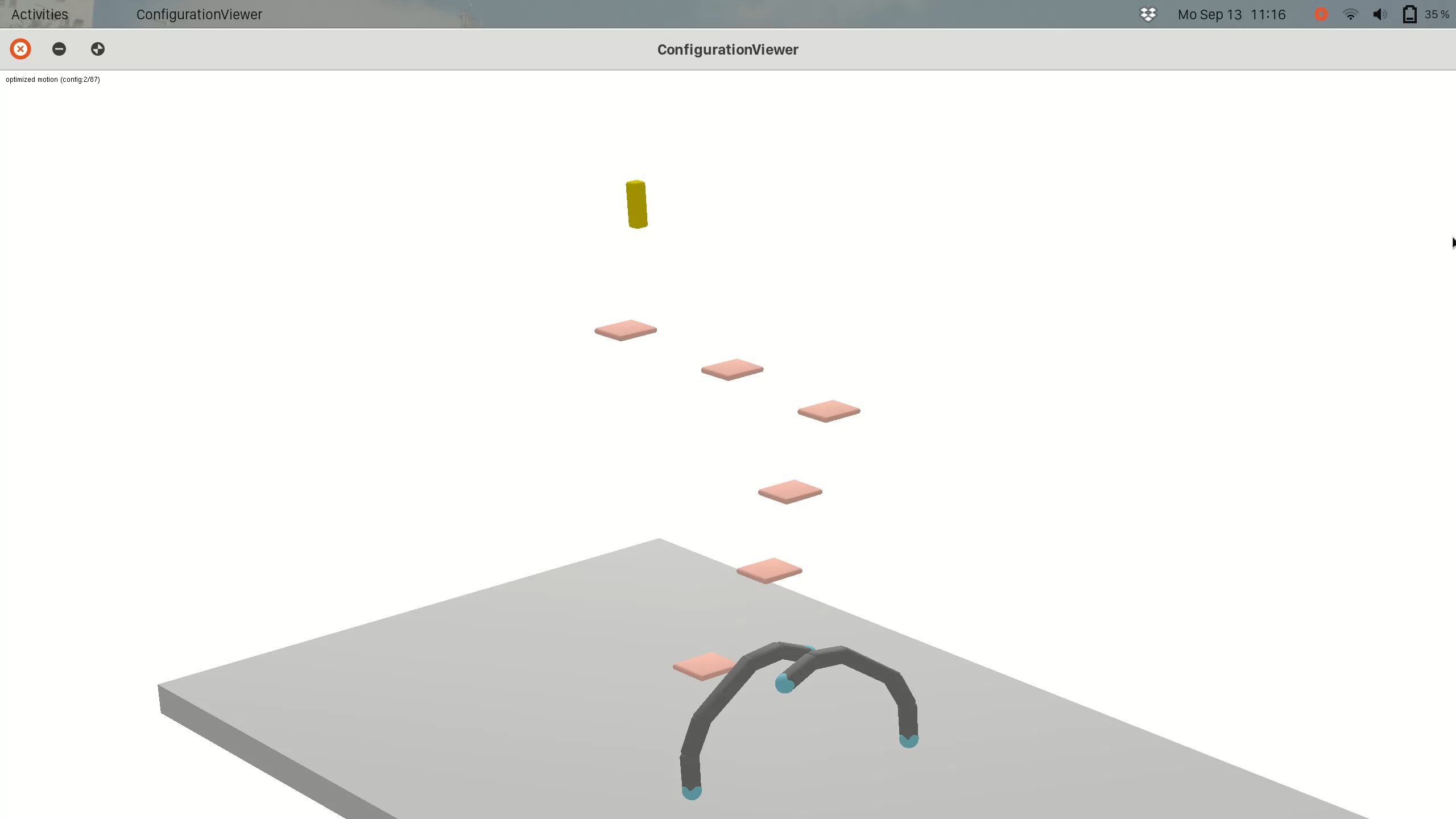}
  \caption{initial config.}
\end{subfigure}%
\hfill
\begin{subfigure}[b]{.17\textwidth}
  \centering
  \includegraphics[width=\textwidth,height=\textwidth,trim=10cm 0cm 10cm 2.75cm,clip]{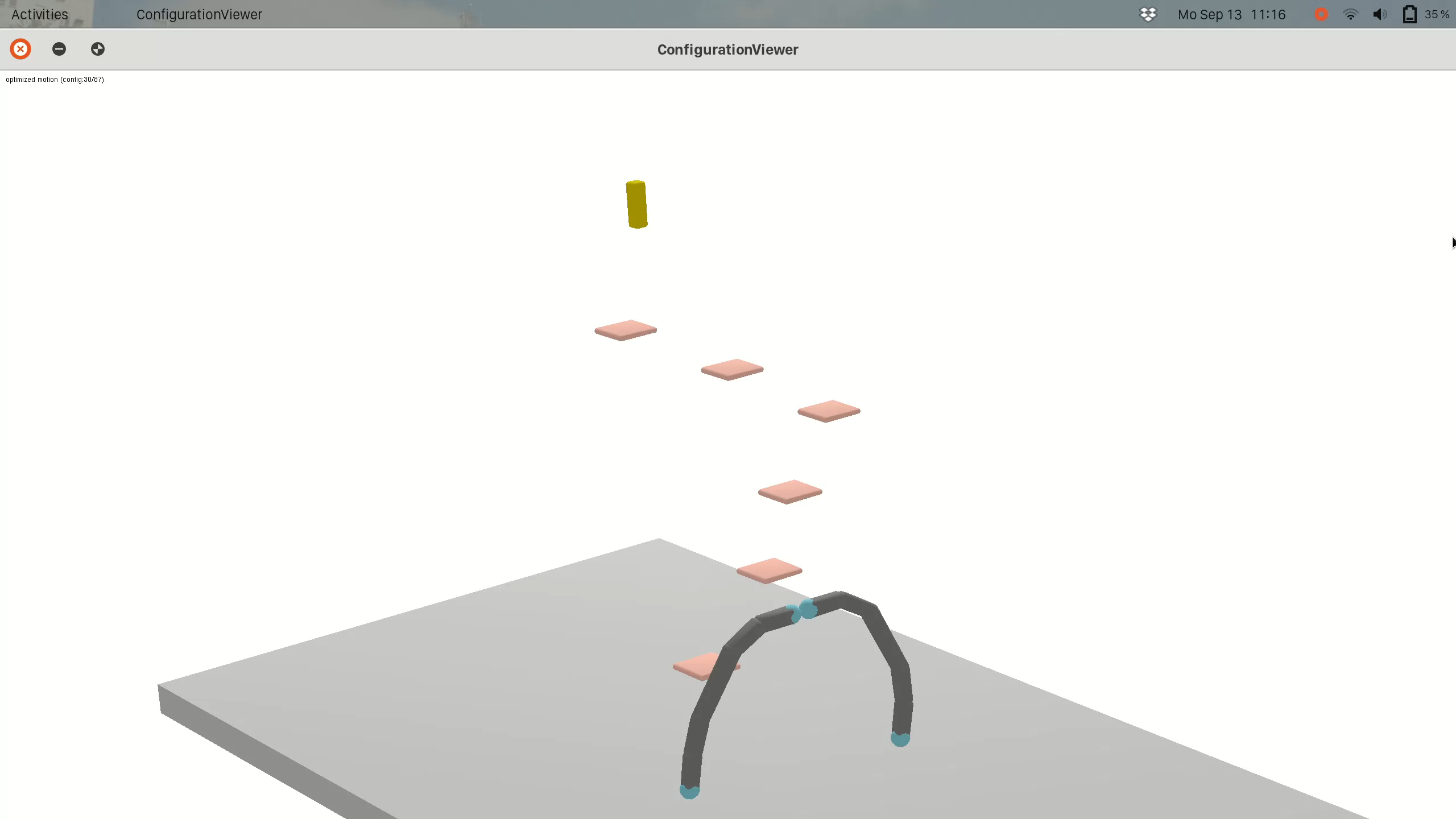}
  \caption{action: connect}
\end{subfigure}%
\hfill%
\begin{subfigure}[b]{.17\textwidth}
  \centering
  \includegraphics[width=\textwidth,height=\textwidth,trim=10cm 0cm 10cm 2.75cm,clip]{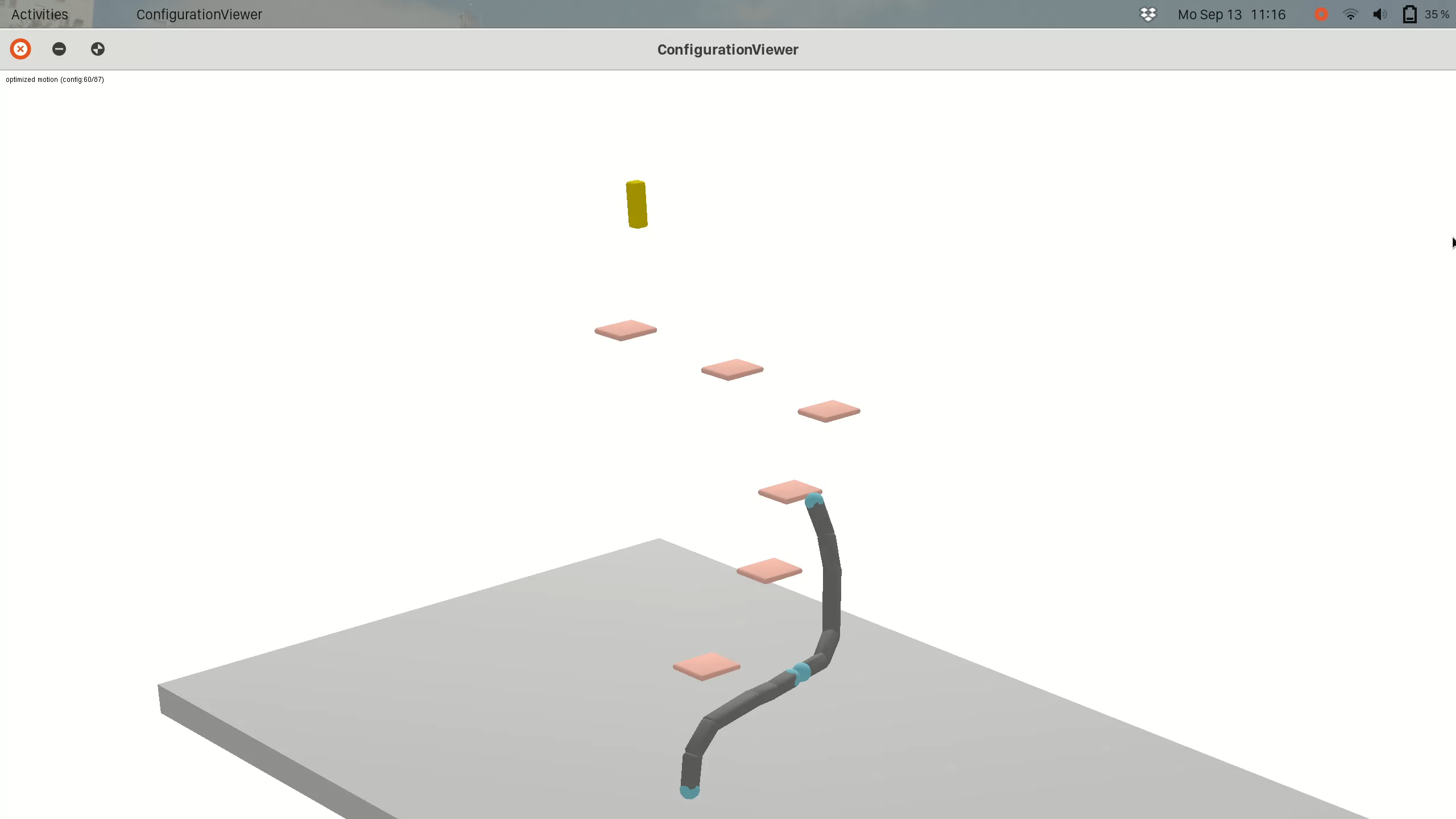}
  \caption{action: stepTogether}
\end{subfigure}%
\hfill
\begin{subfigure}[b]{.17\textwidth}
  \centering
  \includegraphics[width=\textwidth,height=\textwidth,trim=10cm 0cm 10cm 2.75cm,clip]{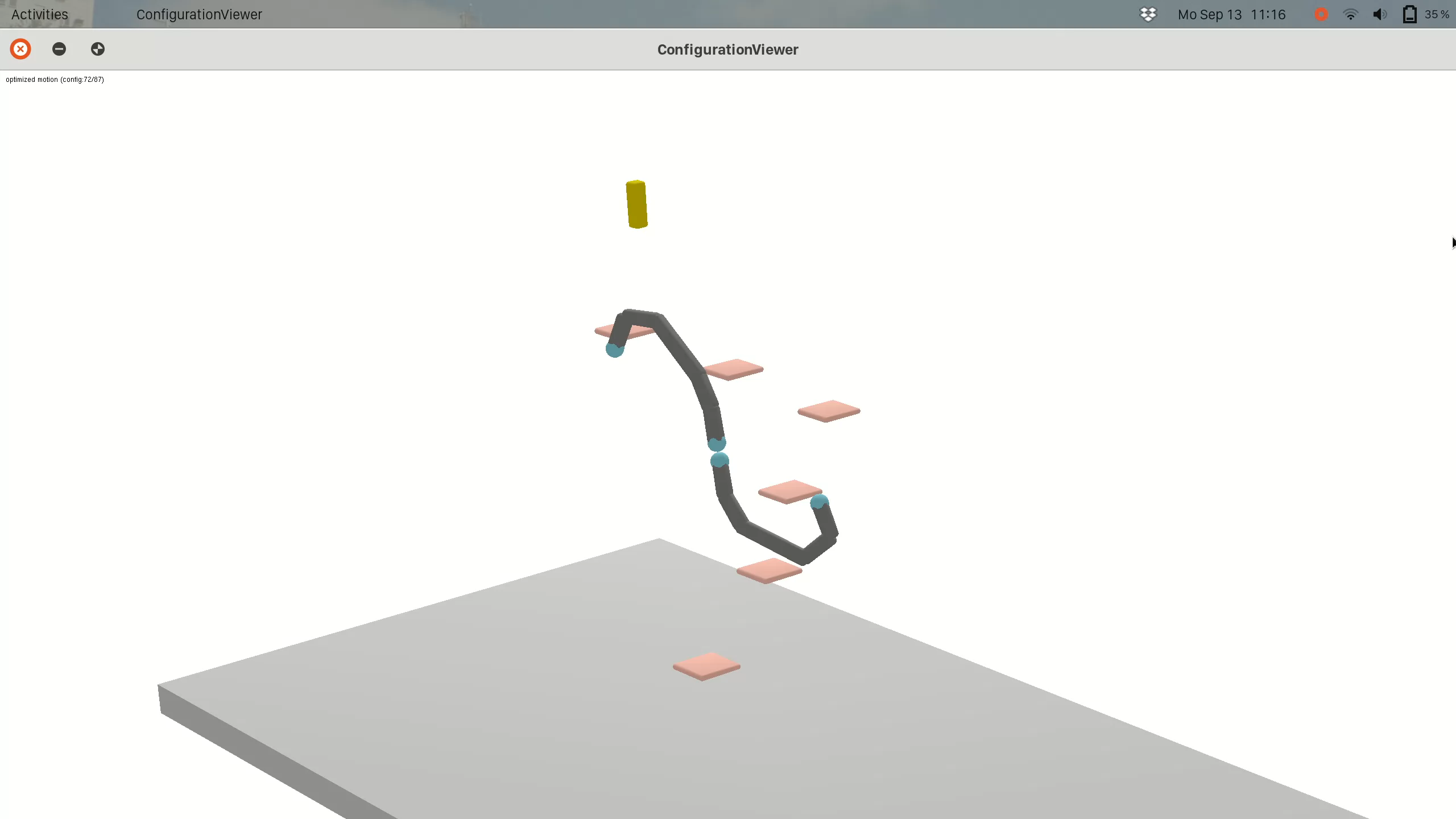}
  \caption{action: stepTogether}
\end{subfigure}%
\hfill
\begin{subfigure}[b]{.17\textwidth}
  \centering
  \includegraphics[width=\textwidth,height=\textwidth,trim=10cm 0cm 10cm 2.5cm,clip]{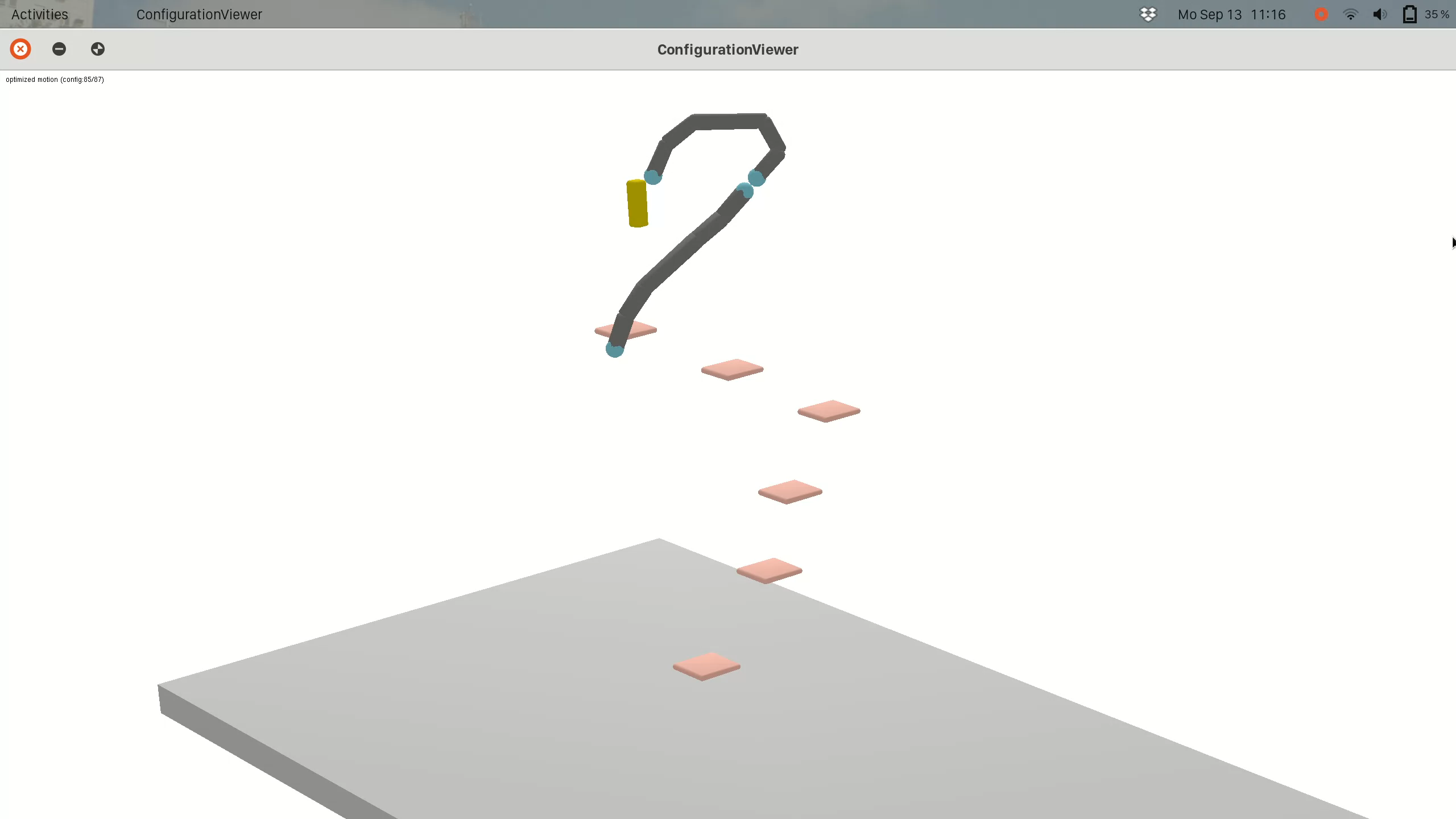}
  \caption{action: pick}
\end{subfigure}%

\caption{The version of the climbing task used in the second experiment.}\label{fig:climb}
\end{figure*}
\vspace{-1pt}

With the guidance of our heuristic framework, LGP can solve manipulation tasks that require very long action sequences (up to 50 actions, see Table \ref{tbl:res}).
In this section, we present a further improvement of this framework, which performs iterative planning.

In the standard version of LGP (Sec. \ref{secBackground}) paths in configuration space (for a fixed sequence of symbolic actions) are optimized jointly, i.e., considering all steps of the manipulation sequences. This strategy does not scale efficiently to very long sequences, as NLP solvers require more iterations to converge, and get often trapped in infeasible local optima, in comparison to shorter manipulation sequences.
To tackle this issue, we propose a receding horizon formulation to optimize paths, that we call RHH-LGP. 
%
Given a full-length candidate sequence of actions to reach the goal, the RHH-LGP path optimization uses only a predefined number of actions $h \in \N$ as a planning horizon and iteratively plans motions for that horizon (between 3 and 10 actions in our experiments).
Once the motions for a horizon are executed, the optimization time window is shifted, and we optimize motions and actions for the subsequent part of the task,  keeping the previous path and actions fixed.
See Alg.~\ref{alg:lgp}, where the integration of the receding horizon formulation is highlighted in blue.
With this receding horizon formulation, RHH-LGP can be applied directly to dynamical environments, where planning and execution run iteratively until the symbolic goal is achieved. If the environment changes during the execution of the current horizon, a new action sequence is computed from the current symbolic and kinematic state. 
This strategy is especially useful when the long action sequence can be decomposed into smaller steps, for example in locomotion tasks. One of the advantages of this approach is that path infeasibility arising from earlier actions in the sequence can be detected efficiently.

\section{Experiments}\label{secExperiments}
\begin{figure}[htbp]
\centering
\begin{subfigure}[b]{.15\textwidth}
  \centering
  \includegraphics[width=\linewidth,trim=6cm 0cm 3cm 2cm,clip]{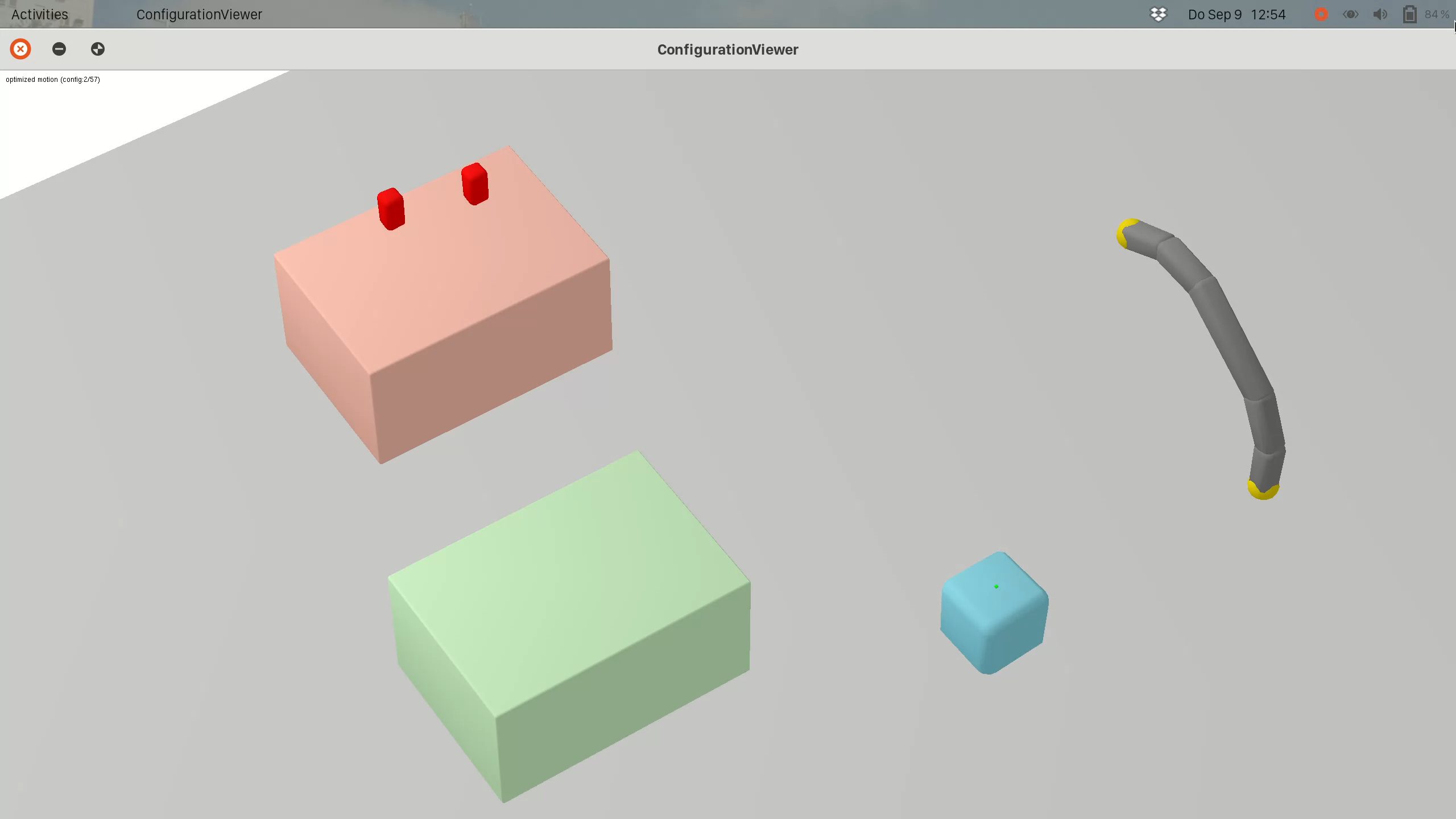}
  \caption{initial config.}\label{fig:pnpA}
\end{subfigure}%
\hfill
\begin{subfigure}[b]{.15\textwidth}
  \centering
  \includegraphics[width=\linewidth,trim=6cm 0cm 3cm 2cm,clip]{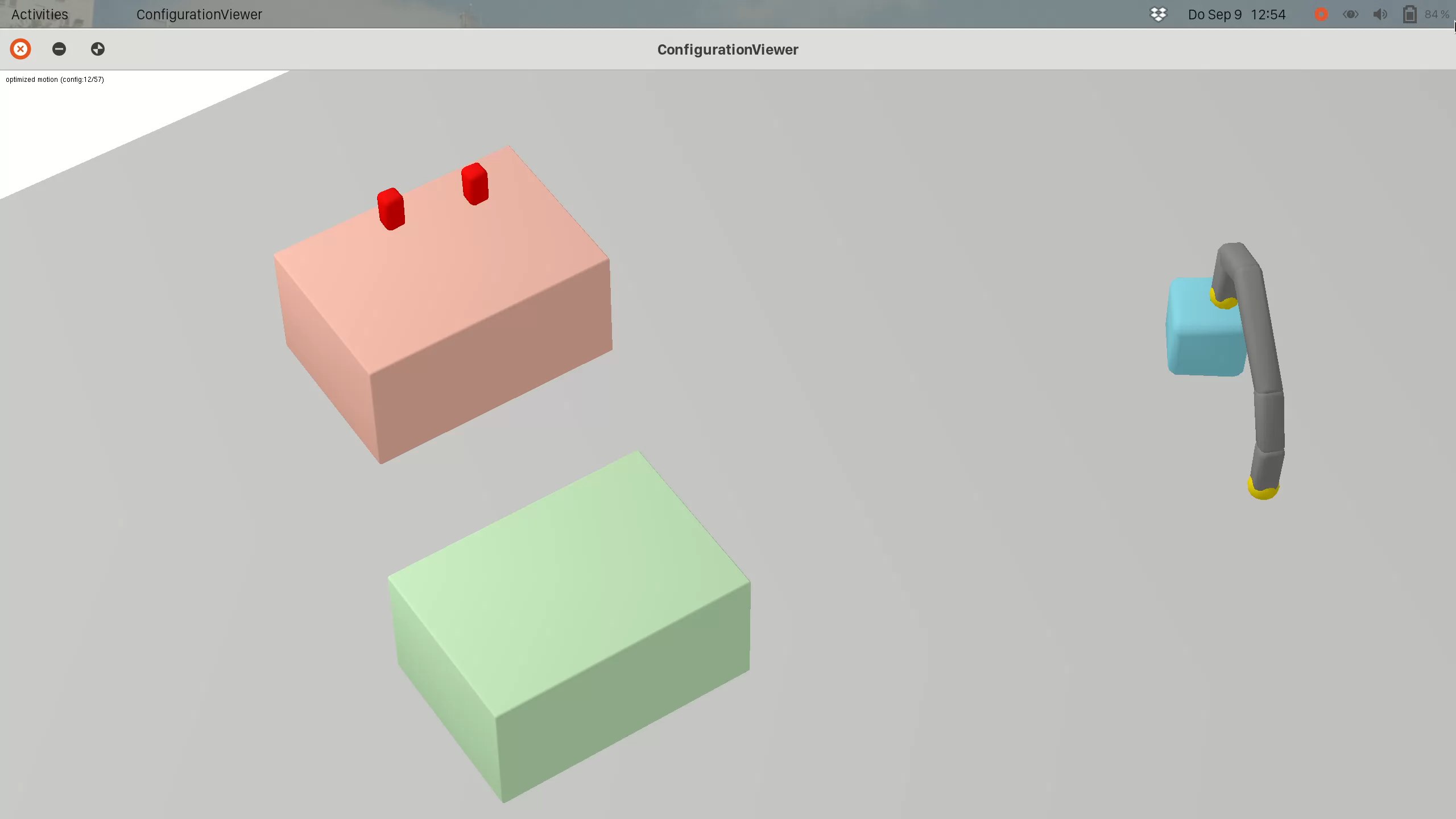}
  \caption{action: connect}
\end{subfigure}%
\hfill%
\begin{subfigure}[b]{.15\textwidth}
  \centering
  \includegraphics[width=\linewidth,trim=6cm 0cm 3cm 2cm,clip]{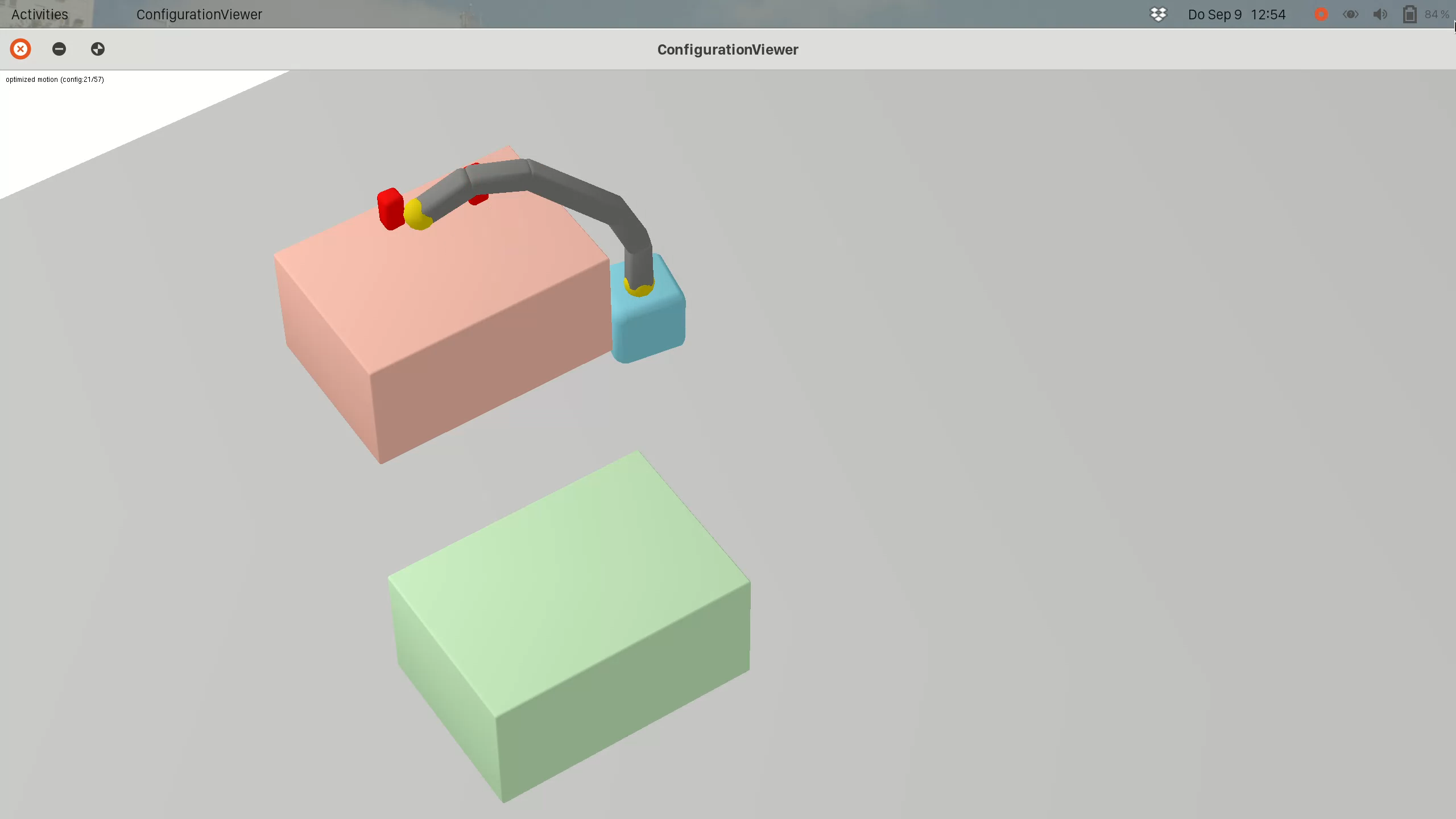}
  \caption{action: pick}
\end{subfigure}%
\\
\begin{subfigure}[b]{.15\textwidth}
  \centering
  \includegraphics[width=\linewidth,trim=6cm 0cm 3cm 2cm,clip]{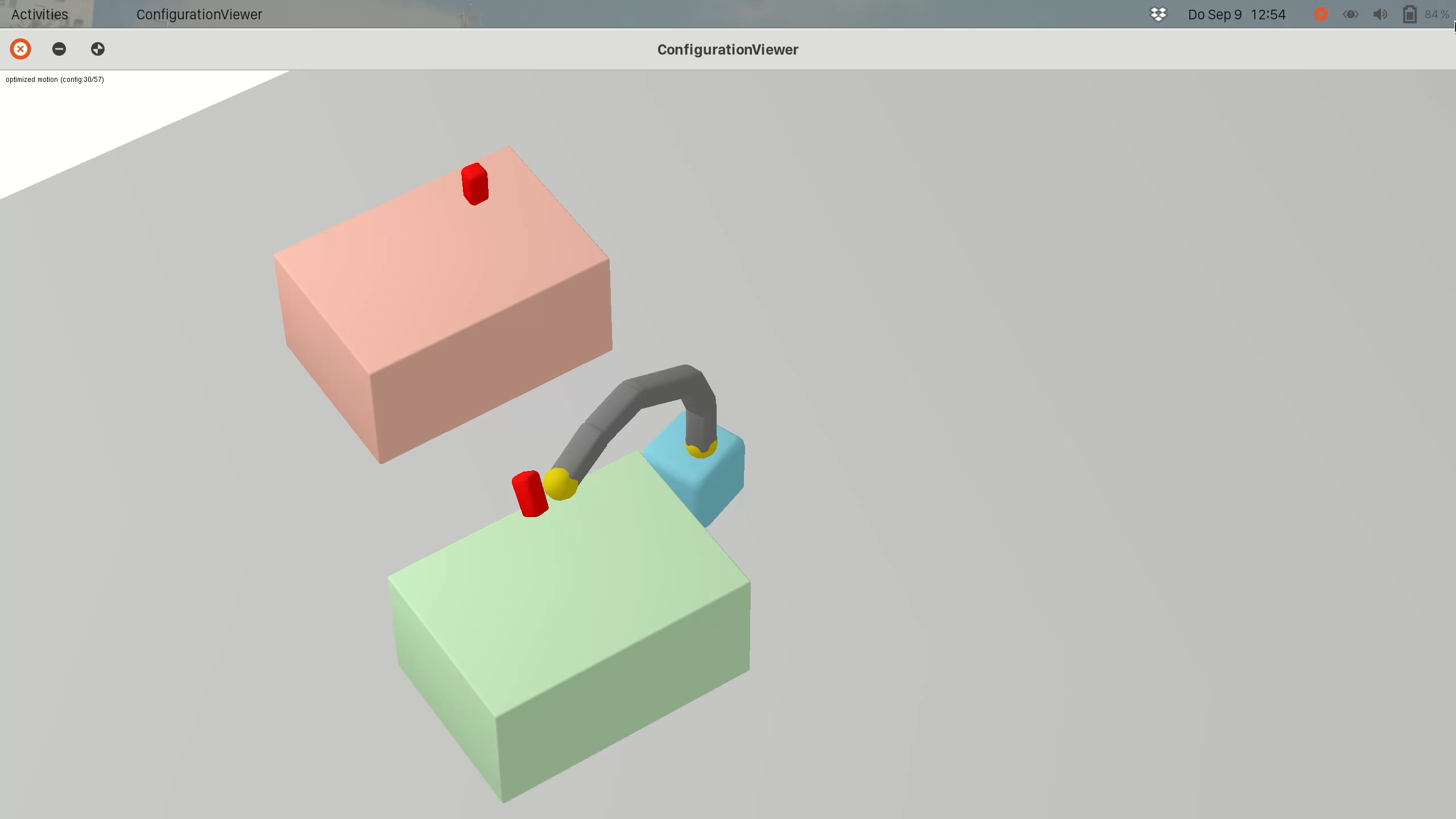}
  \caption{action: place}
\end{subfigure}%
\hfill
\begin{subfigure}[b]{.15\textwidth}
  \centering
  \includegraphics[width=\linewidth,trim=6cm 0cm 3cm 2cm,clip]{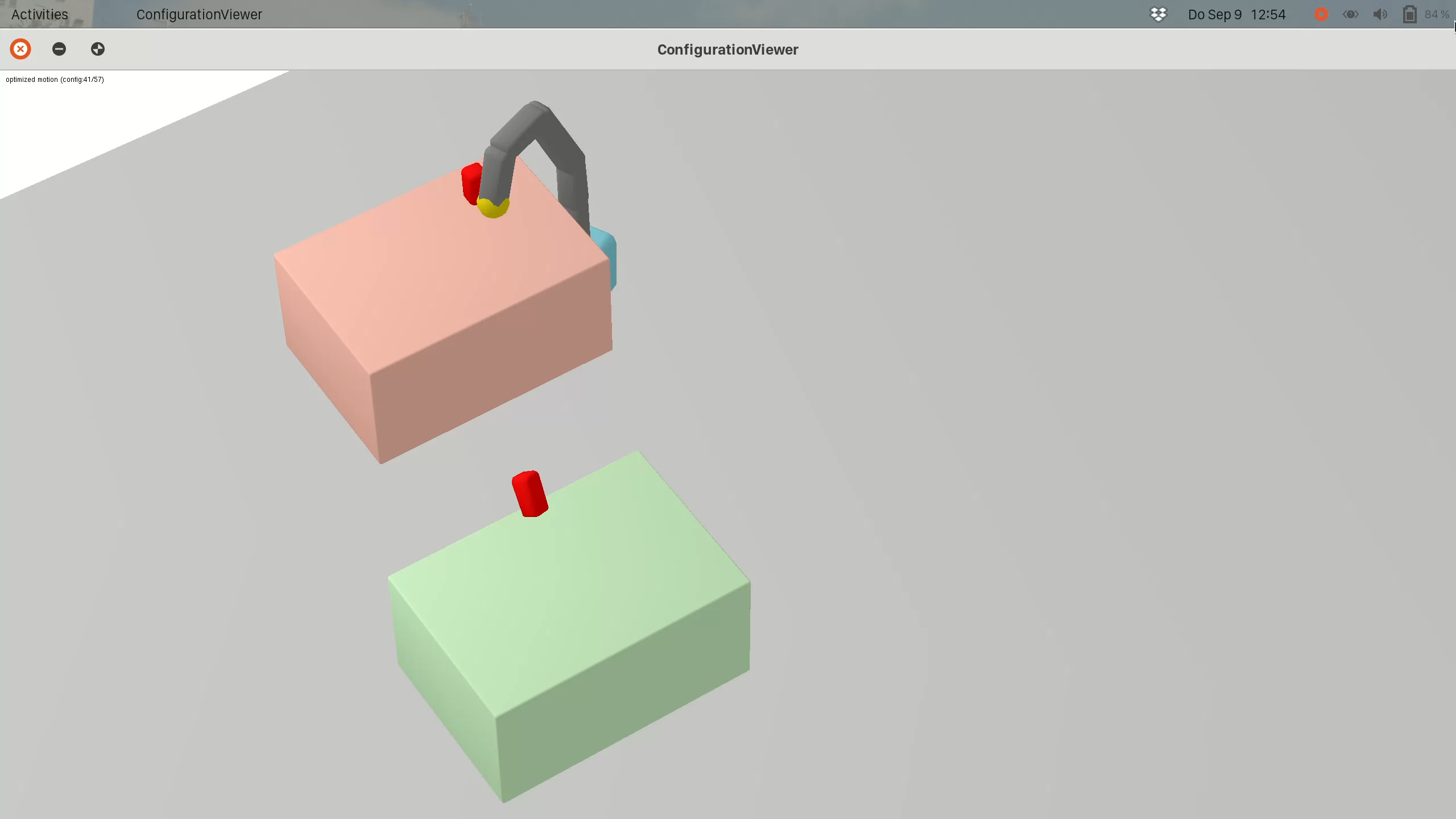}
  \caption{action: pick}
\end{subfigure}%
\hfill
\begin{subfigure}[b]{.15\textwidth}
  \centering
  \includegraphics[width=\linewidth,trim=6cm 0cm 3cm 2cm,clip]{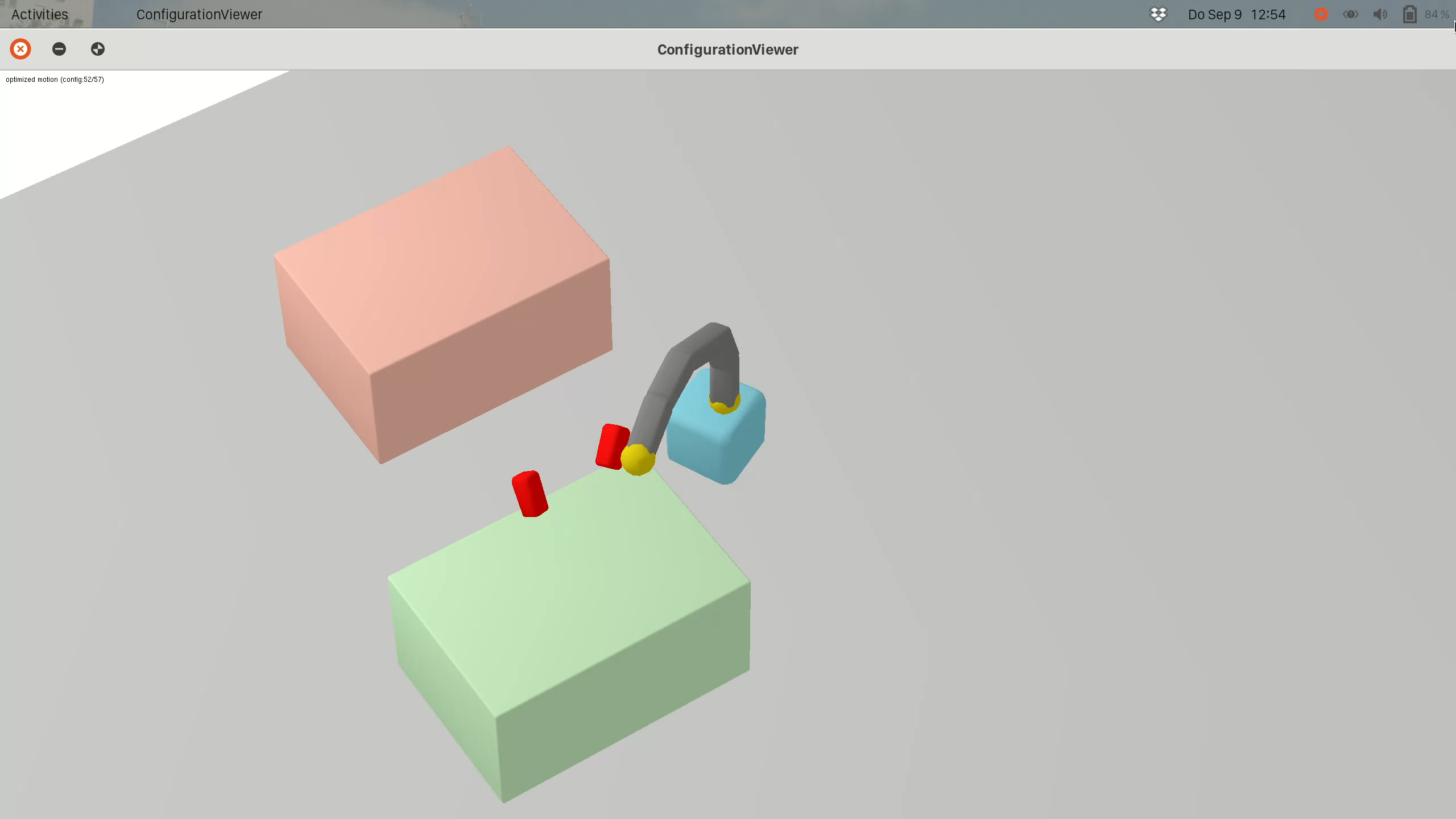}
  \caption{action: place}
\end{subfigure}%
\caption{The modular robot pick-and-place task.}\label{fig:pnp}
\end{figure}

\begin{figure*}[htbp]
\centering
\includegraphics[width=.13\textwidth,trim=7cm 2cm 6cm 3cm,clip]{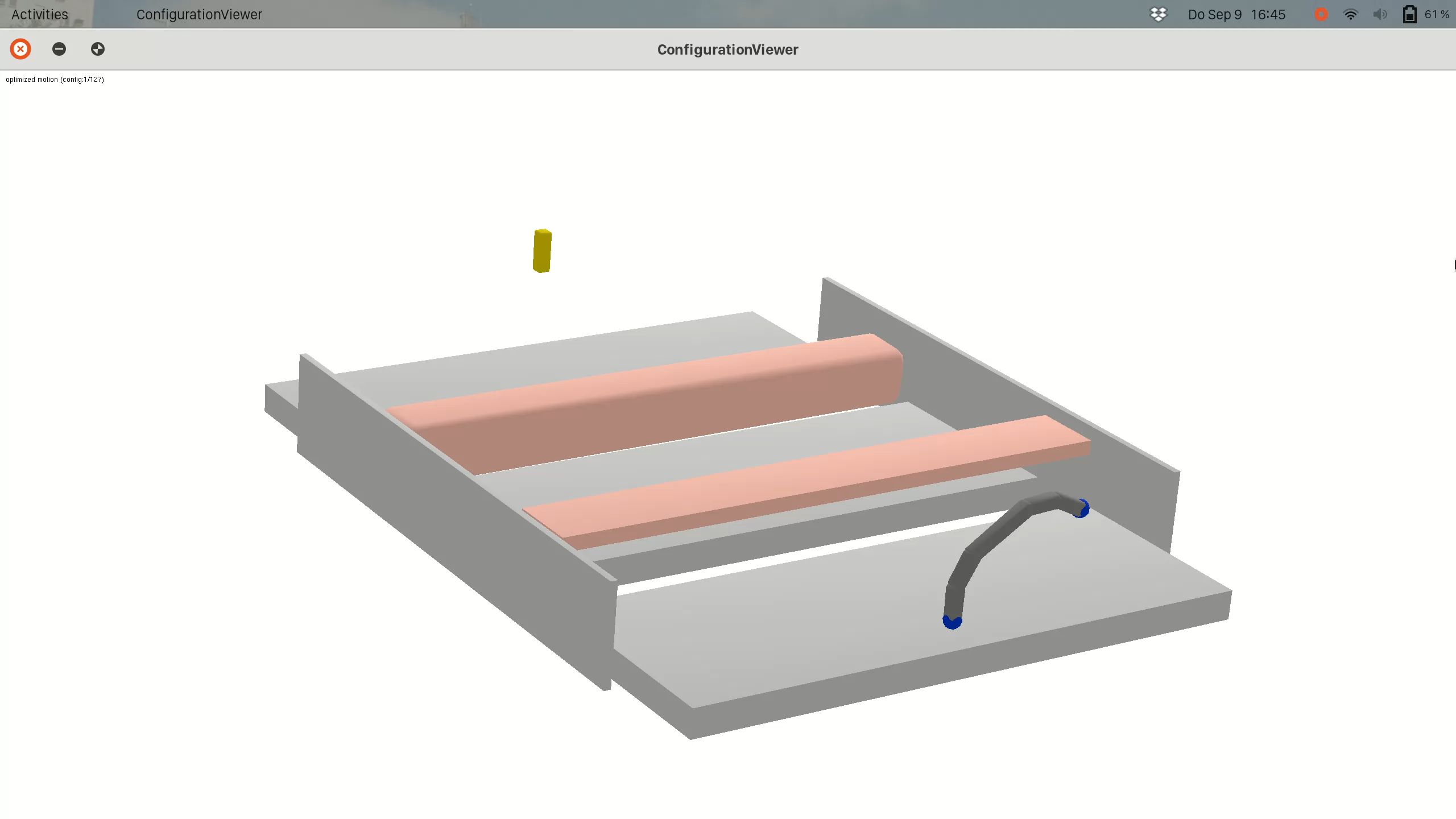}
\includegraphics[width=.13\textwidth,trim=7cm 2cm 6cm 3cm,clip]{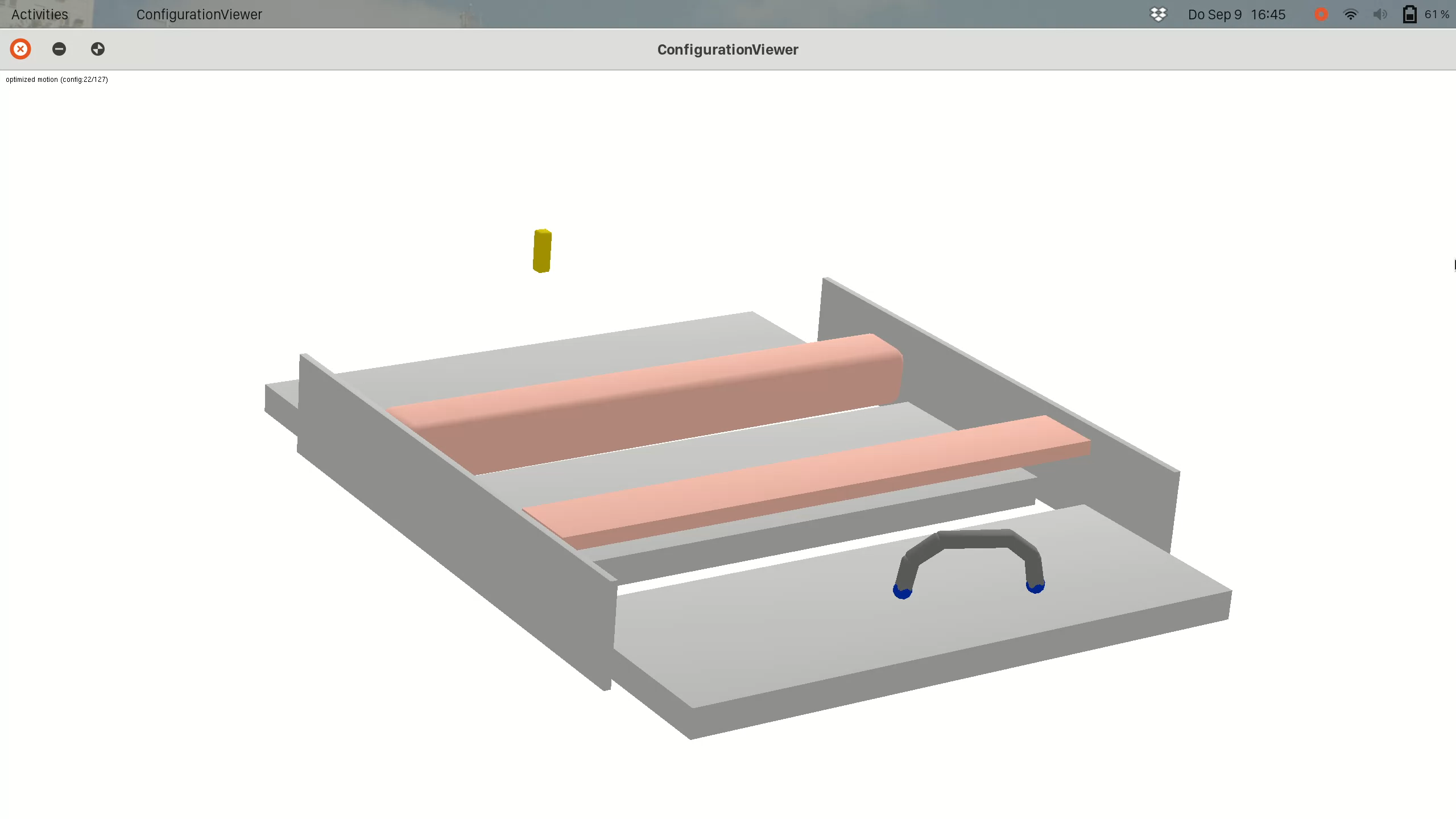}
\includegraphics[width=.13\textwidth,trim=7cm 2cm 6cm 3cm,clip]{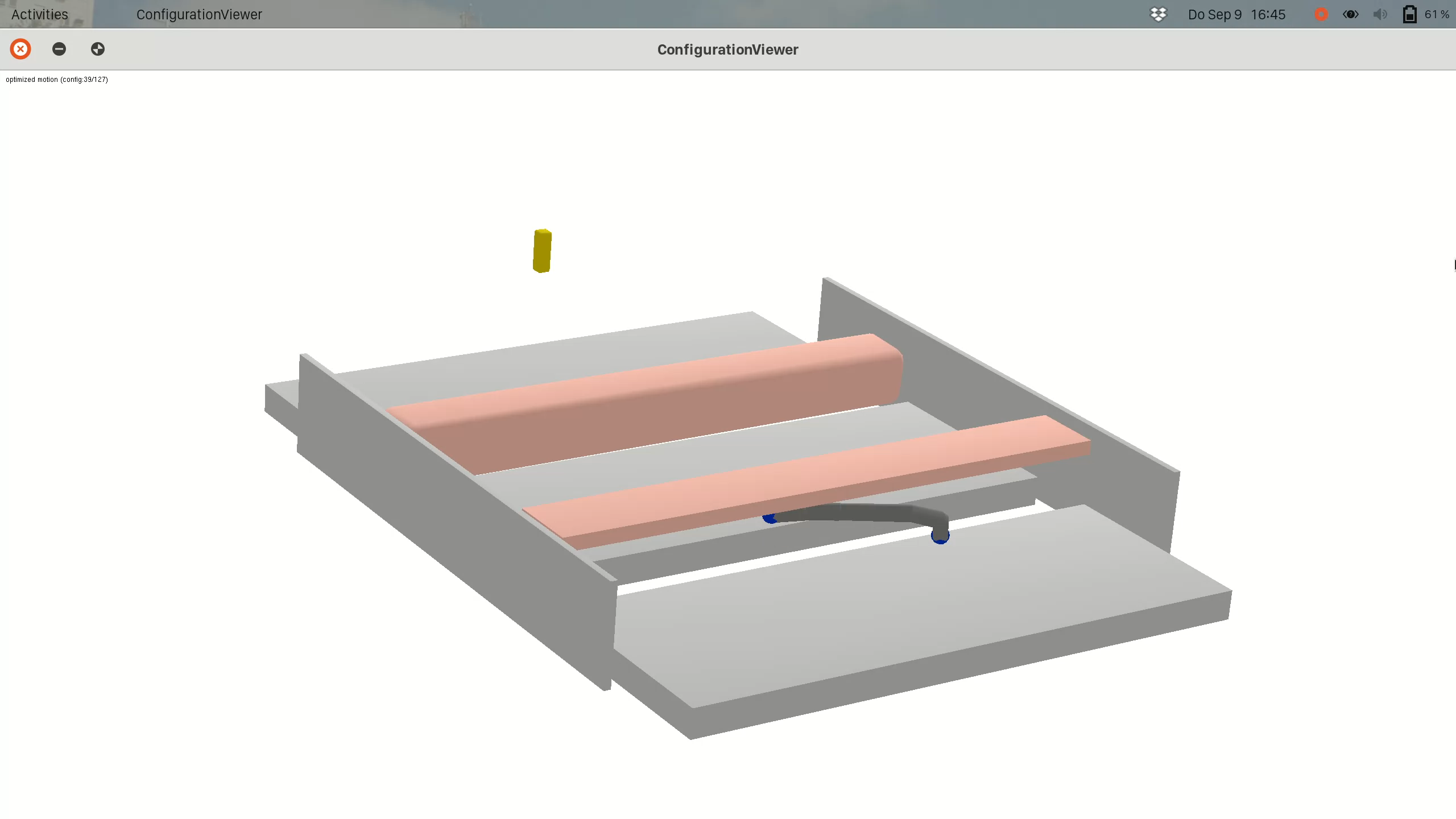}
\includegraphics[width=.13\textwidth,trim=7cm 2cm 6cm 3cm,clip]{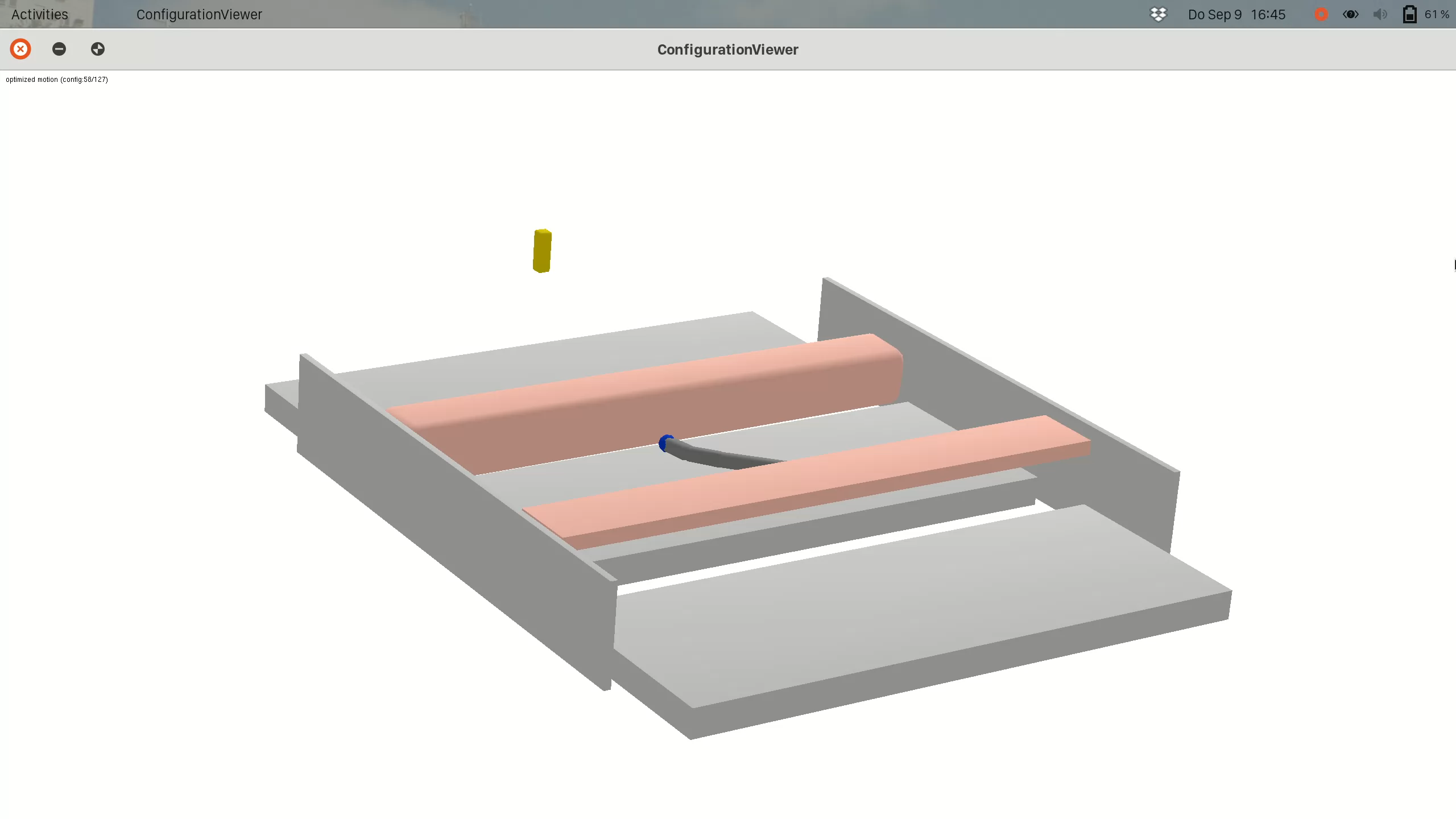}
\includegraphics[width=.13\textwidth,trim=7cm 2cm 6cm 3cm,clip]{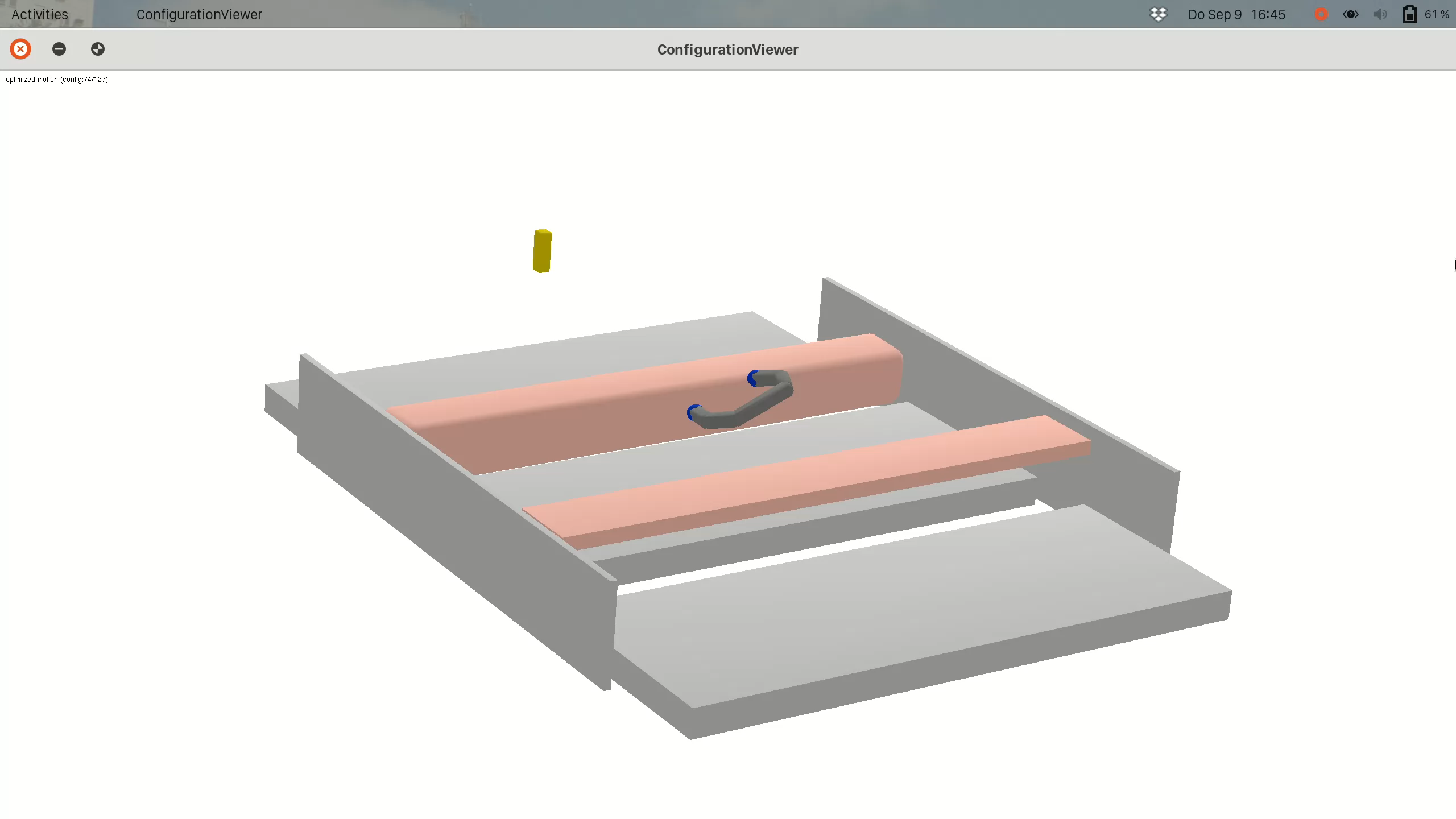}
\includegraphics[width=.13\textwidth,trim=7cm 2cm 6cm 3cm,clip]{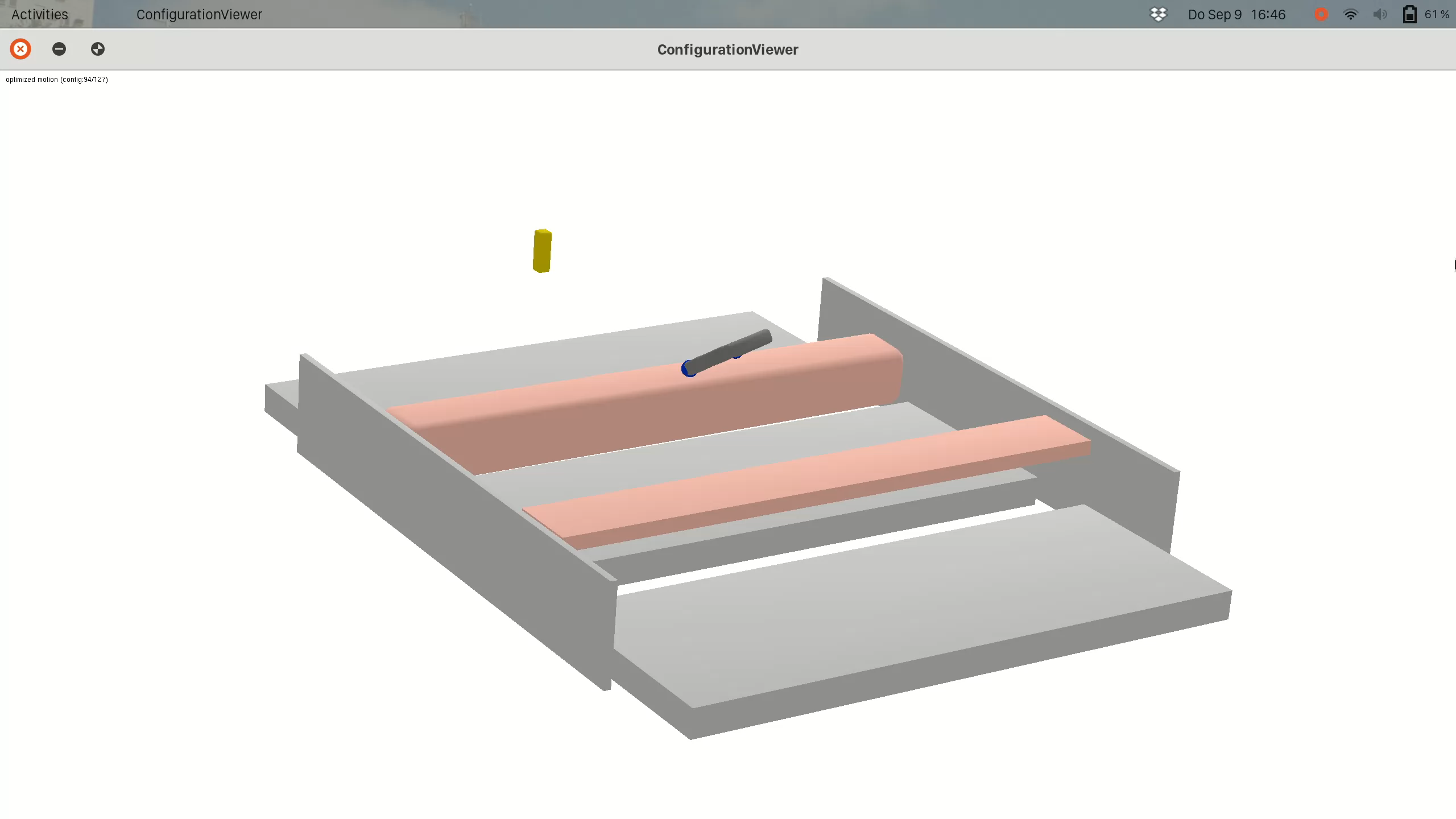}
\includegraphics[width=.13\textwidth,trim=7cm 2cm 6cm 3cm,clip]{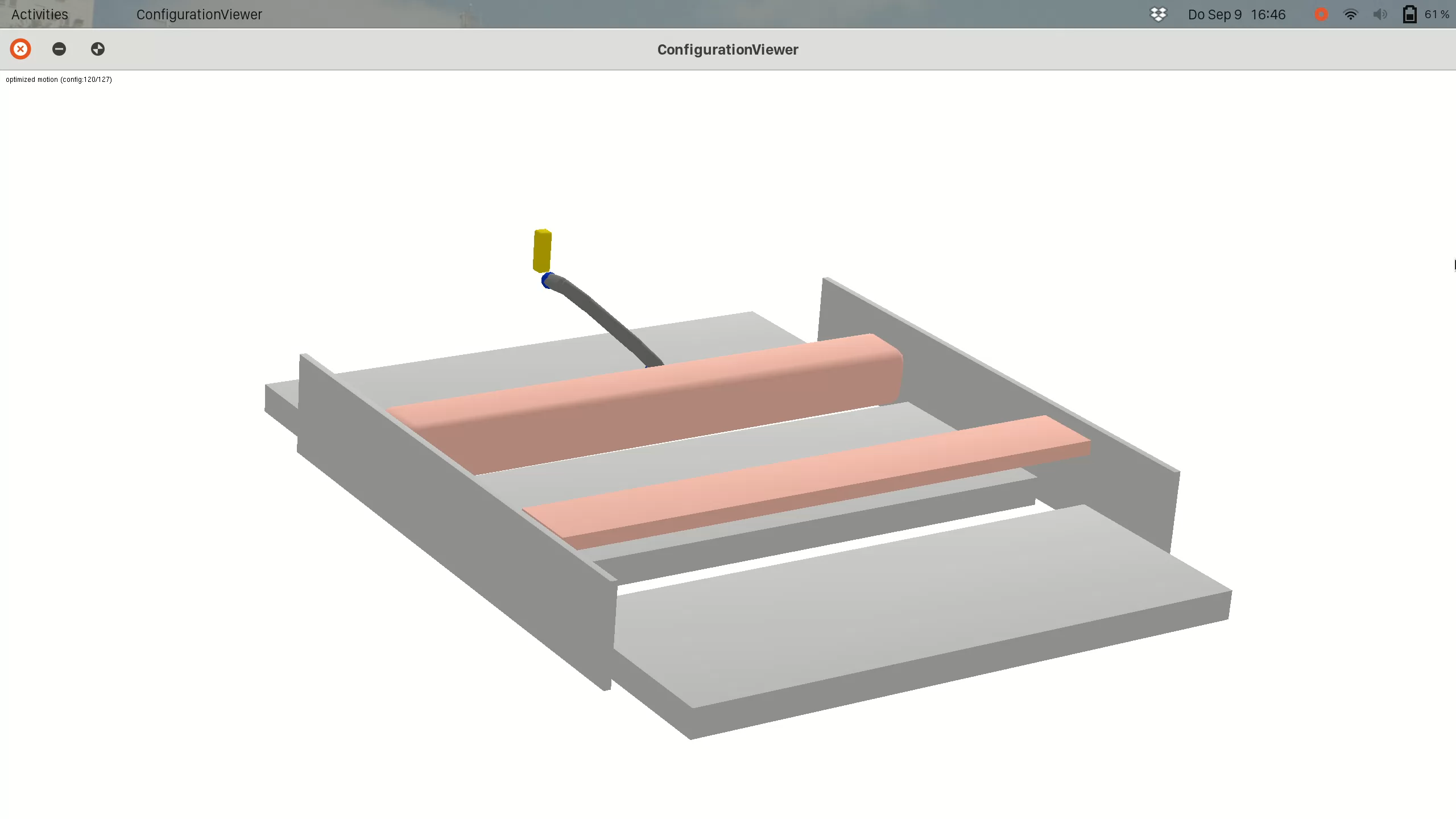}
\caption{The obstacle avoidance task.}\label{fig:obstacle}
\end{figure*}

\begin{table*}[htbp]
\centering
\resizebox{\textwidth}{!}{%
\begin{tabular}{@{}crrrrcrrrrcrrrr@{}} 
\toprule
\multicolumn{1}{c}{$n$} 
& \multicolumn{4}{c}{LGP + Action-specific heuristics} 
& \phantom{abc} 
& \multicolumn{4}{c}{LGP + $\PP_{sequence}$ heuristic} 
& \phantom{abc} 
& \multicolumn{4}{c}{LGP + No Heuristics} \\
\cmidrule{2-5} \cmidrule{7-10} \cmidrule{12-15} 
& Time in s. & \# Tree nodes & \# Expanded & Sol. len. && Time in s. & \# Tree nodes & \# Expanded & Sol. len. && Time in s. & \# Tree nodes & \# Expanded & Sol. len.\\
\midrule
1       & 0.48        & 21      & 3         & 2  && 2.36 & 17 & 1 & 2 && 2.09        & 65     & 8        & 2 \\
2       & 7.00        & 81      & 13        & 7  && 6.23 & 31 & 3 & 2 && 7.91        & 211    & 22       & 2 \\
4       & 13.59       & 190     & 22        & 8  && 127.27 & 225 & 16 & 10 && 313.42 & 7246   & 591      & 5 \\
8       & 11.64       & 309     & 23        & 9  && -- & -- & -- & -- && --          & --     & --       & -- \\
16      & 16.38       & 590     & 26        & 12 && -- & -- & -- & -- && --          & --     & --       & -- \\
32      & 138.03      & 2066    & 47        & 20 && -- & -- & -- & -- && --          & --     & --       & -- \\
64      & 439.01      & 4088    & 53        & 27 && -- & -- & -- & -- && --          & --     & --       & -- \\
\bottomrule
\multicolumn{15}{l}{\footnotesize Note: scenarios without metrics could not be solved at all}\\
\end{tabular}}
\caption{Solving the climbing task for several numbers of stairs $n$ using MBTS and different types of heuristic information.}\label{tbl:exp1}
\end{table*}
We report five different tasks that we used to compare the performance of our LGP solver\footnote{See supp. video for task demonstrations. The source code for the experiments can be found on \href{https://github.com/cornelius-braun/rhh-lgp}{github.com/cornelius-braun/rhh-lgp}.}.
\begin{enumerate}[label=\textit{(\roman*)},leftmargin=*]
    \item \label{task:1} \textit{One crawler climbing to reach a distant target:} In this scenario, the crawler robot has to climb stairs to touch a distant target. This is similar to Fig.~\ref{fig:climb}, but with a single robot only.
    
    \item \label{task:2} \textit{Two crawlers climbing to reach a distant target:} This scenario is similar to the first one, but with two robot modules in the scene.
    This means that a possible solution is that the crawlers connect before climbing the stairs to reach the target, taking bigger steps than a one-module crawler, as illustrated in Fig.~\ref{fig:climb}.
    
    \item \label{task:3} \textit{Climbing with multiple goals:} This task is a variation of the previous task with two targets that have to be reached simultaneously, which makes it a multi-goal task.
    Furthermore, the crawlers initially stand far away from each other and have to walk towards each other in order to connect (Fig.~\ref{fig:complex}).
    
    \item \label{task:6} \textit{Multi-robot pick-and-place task:} Several robotic arms need to transfer multiple objects from table to table in order to place all objects on the goal tray (Fig.~\ref{fig:else}). 
    
    \item \label{task:4} \textit{Modular robot pick-and-place task:} This setting includes one crawler robot and one mobile base robot, initially unconnected (Fig. \ref{fig:pnpA}), and several objects that must be transferred from their initial position to a target tray.
    However, a crawler that holds an object cannot walk, and the solution to this problem involves that the crawler module connects to the mobile base, forming a mobile manipulator (see Fig. \ref{fig:pnp}).

    \item \label{task:5} \textit{Obstacle avoidance task:} A single crawler robot is supposed to navigate towards a target, which is at the end of a series of obstacles (Fig. \ref{fig:obstacle}): a gap in the ground that is covered by a barrier, and a barrier that must be climbed. 
    
\end{enumerate}

Scenario \ref{task:2} is used to benchmark our heuristic framework against two different variations of the standard LGP solver.
\ref{task:1} and \ref{task:2} are used to highlight the benefits of modular robots. 
\ref{task:6}, \ref{task:3} and \ref{task:5} demonstrate the scaling and generality of our approach to more complex scenarios, including multiple robots.
Scenario \ref{task:4} showcases yet another modular robotics capability that involves the dynamic reconfiguration of heterogeneous robot modules.
Lastly, we evaluate the impact of the RHH-LGP solver in all scenarios \ref{task:1} - \ref{task:5}.


\subsection{The impact of the presented heuristics}
To measure the impact of the heuristics on solving TAMP problems, we consider task \ref{task:2} (Fig.~\ref{fig:climb}) and compare the performance of the planner using the action-specific heuristics with the performance of the classic LGP solver and another baseline version of the solver using the result of the $\PP_{sequence}$ bound optimization problem as a heuristic for symbolic search.
In the experiment, we compare the planning performance of the three solvers for several numbers of stairs $n$ in the scene, to assess the scalability of our approach.
The results of the simulation are shown in Table \ref{tbl:exp1}. We highlight that:

\begin{itemize}[leftmargin=*]
    \item The heuristics that we present can be integrated easily into the planning of problems from various domains and improve the planning performance.
    \item For $n\geq8$, only the MBTS solver that used our heuristics can solve the task. This underlines the impact of the presented heuristics on the planning performance.
    \item Using the action-specific heuristics, instances, which are 16 times bigger than the largest problem instance that is tractable for the other two planners, can be solved.
    \item The $\PP_{sequence}$ heuristic solver expands the smallest number of nodes, but this solver is slower in doing so than the heuristics-based solver for $n\geq4$, demonstrating the effectiveness of our approach.
    \item Just as the heuristics that we present, the $\PP_{sequence}$ solver considers geometric information during symbolic planning already, but it uses the motion planner of LGP to estimate the feasibility of the action sequence.
    The results illustrate that the geometry-based heuristics that we present may not be as accurate as the actual motion optimizer, but are much faster and thus more efficient at planning than the optimizer upon each node expansion.
    \item The action-specific heuristics solver enabled to plan action sequences that were up to 3-times longer than the $\PP_{sequence}$ solver and 5-times longer than the standard MBTS solver without any heuristics.
\end{itemize}

\subsection{Iterative planning of the previous tasks}
In this experiment, we plan all tasks \ref{task:1}-\ref{task:5} with the presented RHH-LGP solver that combines action-specific heuristics and iterative planning.
In the following, we will discuss some of the results (Table~\ref{tbl:res}) in more detail:
\begin{itemize}[leftmargin=*]
    \item The RHH-LGP solver can solve more complex tasks than the non-iterative version: Apart from tasks \ref{task:2} and \ref{task:6}, the results show that the RHH-LGP solver can solve the tasks for bigger $n$ than the non-iterative LGP counterpart.
    \item The RHH-LGP solver and the heuristics can solve modular, e.g. task \ref{task:4}, and non-modular robot tasks, e.g. task \ref{task:1}, alike.
    \item Modular robots facilitate planning: for task \ref{task:1}, we find that reaching an object that is 64 stairs high is computationally intractable. On the other hand, if \textit{two} crawlers perform the same task (setting \ref{task:2}), the task is computationally tractable, because the solution requires fewer steps than a possible solution involving a single crawler.
    Furthermore, task \ref{task:3} can only be solved using modular robots, as the initial configuration of the robot modules cannot execute this task because a crawler cannot move while holding an object.
    Hence, our results demonstrate that modular robots can improve TAMP by facilitating symbolic planning and by increasing the number of available actions.
    \item Our approach generalizes to various types of TAMP tasks: We can see that pick-and-place tasks, climbing tasks, and obstacle avoidance tasks can be planned using the presented RHH-LGP algorithm. For each of these tasks, we use different combinations of the same set of action-specific heuristics, which underlines the versatility and generality of the proposed heuristics.
    \item We demonstrate that our solver can handle heterogeneous robot module configurations in task \ref{task:4}.
\end{itemize}
\begin{table*}[htbp]
\centering
\resizebox{\textwidth}{!}{\begin{tabular} {@{}llrrrrcrrrrr@{}} 
\toprule
\multicolumn{1}{c}{Task}
& \multicolumn{1}{c}{$n$} 
& \multicolumn{4}{c}{Non-iterative LGP + action-specific heuristics} 
& \phantom{abc} 
& \multicolumn{5}{c}{RHH-LGP (Iterative LGP + action-specific heuristics)} \\
\cmidrule{3-6} \cmidrule{8-12} && Time in s. & \# Tree nodes & \# Expanded & Sol. len. && Horizon & Time in s. & \# Tree nodes & \# Expanded & Sol. len. \\
\midrule
\multirow{2}{*}{One crawler climbing \ref{task:1}} 
& $32^*$    & 444.26    & 5849  & 171   & 50    && 6  & 85.75    & 5055    & 151 & 33 \\
& $64^*$    & --        & --    & --    & --    && 6  & 255.37   & 13188   & 284 & 47 \\ \hline
\multirow{2}{*}{Two crawlers climbing \ref{task:2}}
& $32^*$    & 138.03    & 2066  & 47    & 19     && 3  & 51.54    & 2341    & 63 & 13 \\ 
& $64^*$    & 439.01    & 4088  & 53    & 27    && 3  & 214.75   & 1083    & 151 & 26  \\ \hline
\multirow{2}{*}{Climbing w. multiple goals \ref{task:3}} 
& $8^{*}$     & 29.48     & 1157  & 77    & 12    && 10 & 50.25    & 1718    & 114 & 13 \\
& $25^{*}$      & --      & --    & --    & --    && 10 & 261.49   & 5824    & 174 & 21 \\ \hline
\multirow{2}{*}{Panda pick-and-place \ref{task:6}} 
& $2^{**}$     & 11.78     & 585  & 95    & 12    && 10 & 7.97    & 637    & 104 & 12 \\
& $4^{**}$      & 30.60      & 5209    & 487    & 24    && 10 & 21.80   & 6637    & 621 & 24 \\ \hline
\multirow{2}{*}{Modular pick-and-place \ref{task:4}} 
& $8^{**}$      & 17.36    & 63    & 16    & 17    && 6  & 8.73     & 131     & 109 & 17 \\
& $16^{**}$     & --       & --    & --    & --    && 4  & 42.82    & 856     & 191 & 33 \\ \hline
\multirow{2}{*}{Obstacle avoidance \ref{task:5}} 
& $2^{***}$      & 58.01    & 211  & 34   & 12     && 6  & 9.66     & 115     & 19  & 9 \\
& $4^{***}$      & --      & --    & --        & -- && 6  & 24.92    & 777     & 77  & 22 \\
\bottomrule
\multicolumn{12}{l}{\footnotesize 
*: $n$ refers to the number of stairs in the scene.
**: $n$ refers to the number of objects that need to be placed.
***: $n$ refers to the the number of obstacles.}\\
\multicolumn{10}{l}{\footnotesize Note: scenarios without metrics could not be solved at all}\\
\end{tabular}}
\caption{Comparing the performance of the iterative and non-iterative MBTS solvers using heuristics for various tasks.}\label{tbl:res}
\end{table*}

\section{Discussion}\label{secDiscussion}
Our work can be seen as a first step towards fully autonomous modular robots that can solve a variety of complex and long-horizon tasks.
In this light, there are some aspects of our work that we would like to discuss in more detail.
%
In the experiments, it is demonstrated that the RHH-LGP algorithm yields an order of magnitude improvement in planning time in a variety of scenarios.
This highlights the flexibility and effectiveness of our planner.
Thanks to this informed and iterative version of the LGP formulation, long-horizon tasks (requiring up to 50 symbolic actions) can be solved effectively.
In addition, we demonstrate that tasks in an environment with multiple robots that can connect to form a new entity can be planned.
The experiments demonstrate that when the search space increases, which is usually unavoidable in multi-agent scenarios, using a planner that incorporates geometric knowledge during symbolic planning is crucial.

Still, it remains to be noted that the reachability check that we use is simple:
The approximation of the robot position that we make during symbolic planning is based on the objects that the robots are standing on, assuming that these objects do not move.
Even though this assumption holds for the scenarios we investigated, in general, it does not necessarily hold for all manipulation tasks.

The horizon length of the RHH-LGP solver is an important hyper-parameter of our algorithm.
In the environments we investigated, choosing a suitable horizon length turns out to be sufficient to prevent the receding horizon strategy approach from inducing undesirable infeasibility, i.e., fixing the current path does not result in infeasible paths for future actions.

The issue of fixing previous motions could be solved by adding backtracking: our algorithm could naturally include a backtracking mechanism that restores to full joint optimization if necessary. In essence, if a motion for subsequence $\{a_i,\ldots,a_{i+h}\}$ is not feasible, we iteratively increase the time window backwards and optimize for $\{a_{i-h},\ldots, a_i,\ldots,a_{i+h}\}$.

\section{Conclusion}\label{secConclusion}

There exist many TAMP approaches, which mostly differ in their integration of the logical decision-making process with the motion planning component and/or whether they use sampling-~or~optimization-based motion planning methods.
In this work, we extended an optimization-based approach called LGP. 
Even though prior studies have demonstrated the effectiveness of LGP on sequential physical manipulation problems, these tasks mainly involve short-horizon pick-and-place scenarios.
This work investigates improving this state-of-the-art TAMP solver to address longer horizon problems, which also comprise a wider variety of tasks that have not been examined previously.

By introducing simple yet generic geometry-based heuristics along with an iterative receding horizon approach, which we call RHH-LGP, the performance of logic-geometric programming improves remarkably.
This improvement allows solving significantly longer horizon tasks than previously shown.
In our experiments, we investigated a diverse set of problems, ranging from pick-and-place tasks to scenarios in which heterogeneous robots need to dynamically form novel kinematic structures.
Our results demonstrate that the proposed algorithm solves these tasks effectively, while the original LGP solver fails to find solutions for the majority of these problems.
By validating our approach over a variety of scenarios, we not only showcase the efficacy of our planner but also its flexibility to many different tasks.
Our work highlights that LGP-based TAMP approaches, when enhanced by an iterative and heuristics augmented formulation, have the potential to be applied to different types of long-horizon problems. 

\\




\bibliographystyle{IEEEtran}
\typeout{} 
\bibliography{IEEEabrv,mybibfile}


\end{document}